\documentclass{article} 

\usepackage{longtable}

\usepackage{booktabs}
\usepackage[load-configurations=version-1]{siunitx} 

\usepackage[utf8]{inputenc}       
\usepackage[T1]{fontenc}          
\usepackage{lmodern}              
\usepackage{hyperref}             
\usepackage{graphicx}             
\usepackage{authblk}              
\usepackage{url}                  

\hypersetup{
    colorlinks=true,
    linkcolor=blue!50!black,   
    citecolor=blue!50!black,   
    urlcolor=blue!50!black     
}
      
\usepackage{arxiv}
\usepackage{fancyhdr}

\usepackage[utf8]{inputenc}
\usepackage[T1]{fontenc}
\usepackage{times}

\usepackage{amsmath}
\usepackage{amssymb}
\usepackage{amsfonts}
\usepackage{amsthm}
\usepackage{mathtools}
\usepackage{nicefrac}
\usepackage{latexsym}
\usepackage{eucal}

\usepackage{graphicx}
\usepackage{wrapfig}
\usepackage{subfig}
\usepackage{float}

\usepackage{booktabs}
\usepackage{tabularx}
\usepackage{multirow}
\usepackage{multicol}
\usepackage{makecell}
\usepackage{cellspace}
\usepackage{siunitx}

\usepackage[table, svgnames, dvipsnames]{xcolor}

\usepackage[justification=centering]{caption}

\usepackage{enumitem}

\usepackage[noend]{algpseudocode}
\usepackage{algorithm}

\usepackage{soul}
\usepackage{microtype}
\usepackage{blindtext}
\usepackage{lipsum}

\usepackage{natbib}
\usepackage{doi}
\usepackage{url}
\usepackage{cleveref}

\usepackage{acronym}
\usepackage{glossaries}

\usepackage[misc]{ifsym}
\usepackage{authblk}
\usepackage[switch]{lineno}

\usepackage{amsmath}
\usepackage{amssymb}
\usepackage{amsthm}

\usepackage{graphicx}
\usepackage{booktabs}
\usepackage{tabularx}
\usepackage{multirow}
\usepackage{multicol}

\usepackage{enumitem}
\usepackage{latexsym}
\usepackage{eucal}

\usepackage{soul}
\usepackage{color}
\usepackage{url}

\usepackage{algorithm}
\usepackage{algpseudocode}

\usepackage[switch]{lineno}

\usepackage{acronym}

\newtheorem{definition}{Definition}
\newtheorem{requirement}{Requirement}

\title{Tabular Foundation Models Can Learn\\Association Rules}

\acrodef{DL}{Deep Learning}
\acrodef{ML}{Machine Learning}
\acrodef{CNF}{Conjunctive Normal Form}
\acrodef{ARM}{Association Rule Mining}
\acrodef{NARM}{Numerical Association Rule Mining}
\acrodef{ANN}{Artificial Neural Network}
\acrodef{IoT}{Internet of Things}
\acrodef{PFN}{Prior-Data Fitted Network}
\acrodef{SCM}{Structural Causal Model}
\acrodef{TFM}{Tabular Foundation Model}

\newbox{\orcid}\sbox{\orcid}{\includegraphics[scale=0.06]{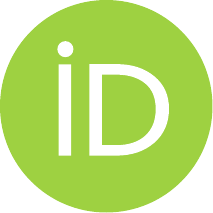}} 

\author[1,2]{%
    \href{https://orcid.org/0000-0003-2710-7951}{\usebox{\orcid}}%
    \hspace{1mm}Erkan Karabulut\thanks{Corresponding author: \texttt{e.karabulut@uva.nl}.}%
}
\author[2]{%
    \href{https://orcid.org/0000-0002-5357-3705}{\usebox{\orcid}}%
    \hspace{1mm}Daniel Daza%
}
\author[1]{%
    \href{https://orcid.org/0000-0003-0183-6910}{\usebox{\orcid}}%
    \hspace{1mm}Paul Groth%
}
\author[3]{%
    \href{https://orcid.org/0000-0003-4591-9646}{\usebox{\orcid}}%
    \hspace{1mm}Martijn C. Schut%
}
\author[1]{%
    \href{https://orcid.org/0000-0001-7054-3770}{\usebox{\orcid}}%
    \hspace{1mm}Victoria Degeler%
}

\affil[1]{University of Amsterdam, The Netherlands}
\affil[2]{Vrije Universiteit Amsterdam, The Netherlands}
\affil[3]{Amsterdam UMC, location VUmc, Vrije Universiteit Amsterdam, the Netherlands}

\begin{document}

\maketitle

\begin{abstract}
Association Rule Mining (ARM) is a fundamental task for knowledge discovery in tabular data and is widely used in high-stakes decision-making. Classical ARM methods rely on frequent itemset mining, leading to rule explosion and poor scalability, while recent neural approaches mitigate these issues but suffer from degraded performance in low-data regimes.
Tabular foundation models (TFMs), pretrained on diverse tabular data with strong in-context generalization, provide a basis for addressing these limitations. We introduce a model-agnostic association rule learning framework that extracts association rules from any conditional probabilistic model over tabular data, enabling us to leverage TFMs. We then introduce \emph{TabProbe}, an instantiation of our framework that utilizes TFMs as conditional probability estimators to learn association rules \emph{out-of-the-box} without frequent itemset mining. We evaluate our approach on tabular datasets of varying sizes based on standard ARM rule quality metrics and downstream classification performance. The results show that TFMs consistently produce concise, high-quality association rules with strong predictive performance and remain robust in low-data settings without task-specific training. Source code is available at \href{https://github.com/DiTEC-project/tabprobe}{https://github.com/DiTEC-project/tabprobe}. 
\end{abstract}

\noindent{\small \textbf{Note:} An extended version of this preprint, entitled \textit{Model-Agnostic Association Rule Learning with Tabular Foundation Models}, is published in TMLR: \url{https://openreview.net/pdf?id=Y70lnNnA6T}.\par}

\section{Introduction}

\ac{ARM} is an unsupervised knowledge discovery task that aims to find associations between features in a given dataset, such as tabular data, in the form of logical implications~\citep{agrawal1994fast}. It has been widely used across numerous domains for knowledge discovery~\citep{kaushik2023numerical,luna2019ARMsurvey} and high-stakes decision-making~\citep{rudinnature} as part of interpretable \ac{ML} models~\citep{corels}.

The ARM literature has been dominated by algorithmic methods that suffer from rule explosion and high computational overhead~\citep{moens2013frequent} without effective search space reduction~\citep{srikant1997mining,baralis2012generalized,yin2022constraint,fournier2012mining,nguyen2018etarm,zaki2002charm}. Such methods typically operate in two phases: mining frequent itemsets, and then generating association rules using the frequent itemsets.

Recent neurosymbolic methods, such as Aerial+~\citep{aerial_plus}, utilize autoencoders~\citep{bank2023autoencoders} to learn compact representations and extract small rule sets, overcoming rule explosion. However, such methods perform poorly on small datasets~\citep{liu2017deep}, leading to inaccurate rules that do not reflect true patterns~\citep{aerial_tabpfn}. This is particularly problematic as small tabular data is common in many domains, including biomedicine, such as rare disease datasets~\citep{rare_diseases}. 

To address this, we propose a model-agnostic framework that enables any existing \textit{\ac{TFM}} with a probabilistic classification objective \textit{out-of-the-box} (that is, without requiring additional training) to learn association rules without frequent itemset mining. TFMs~\citep{tabpfn-nature} have demonstrated strong performance as classifiers for small tabular data by pretraining on large-scale synthetic~\citep{qu2025tabicl} or real~\citep{spinaci2025contexttab} tabular data and transferring learned representations to unseen datasets. The training of TFMs is often performed using in-context learning~\citep{pfns}. They consistently outperform traditional tabular methods. Despite their growing success, TFMs have not yet been directly utilized for ARM.

\begin{table}[!t]
    \centering
    \caption{\textbf{Sample clinical dataset and association rules.} The table shows sample patient records from a hepatitis dataset with a target outcome. Listed association rules are extracted from TabPFN using our approach and capture interpretable patterns relating symptoms and treatments to patient survival.}
    \vspace{5pt}
    \label{tab:hepatitis_rules}
    \begin{tabular}{lccccc}
        \toprule
        \textbf{Age} & \textbf{Sex} & \textbf{Steroid} & \textbf{Antiviral} & $\cdots$ & \textbf{Class} \\
        \midrule
        30 & female & no & yes & $\cdots$ & LIVE \\
        50 & male & no & yes & $\cdots$ & LIVE \\
        78 & male & yes & yes & $\cdots$ & LIVE \\
        39 & male & yes & no & $\cdots$ & DIE \\
        \midrule
        \textbf{Rule 1} & \multicolumn{5}{c}{Antiviral = yes $\wedge$ Malaise = yes $\rightarrow$ Class = LIVE} \\
        \textbf{Rule 2} & \multicolumn{5}{c}{Malaise = yes $\wedge$ Anorexia = yes $\rightarrow$ Class = LIVE} \\
        \bottomrule
    \end{tabular}
\end{table}

Our model-agnostic framework extracts association rules from any conditional probabilistic model trained on tabular data. We instantiate our framework with TFMs by leveraging their in-context learning scheme to develop \textit{TabProbe}: a probing-like rule extraction algorithm. A table is passed as a context to a TFM together with a probing matrix that contains a set of data points with marked features. If a feature set $\mathcal{A}$ leads to a high probability prediction of another set $\mathcal{C}$ such that $P(\mathcal{C}|\mathcal{A}) \ge \tau$ for a given threshold $\tau$, then we conclude with the rule $\mathcal{A} \rightarrow \mathcal{C}$. Samples from the clinical Hepatitis~\citep{Hepatitis} dataset, together with association rules learned with TabPFN~\citep{tabpfn-nature} using TabProbe, are shown in Table \ref{tab:hepatitis_rules}.
    
The main \textit{contributions} of this paper are as follows: 

\begin{enumerate}
    \item We propose a model-agnostic framework for extracting association rules from probabilistic models by conditional feature validation and prediction (Section~\ref{sec:framework}).
    \item We provide two instantiations of the framework; \textit{TabProbe}, a novel algorithm for association rule learning with TFMs, and one using reconstruction models such as autoencoders~\citep{bank2023autoencoders} (Section~\ref{sec:framework-init}).
    \item We show that TFMs can learn \textit{out-of-the-box} a concise set of high-quality association rules with equal or higher predictive performance than state-of-the-art ARM methods and remain robust in low-data regimes. This is shown through a comprehensive evaluation on small ($\sim$100 rows) and larger tabular datasets ($\geq 1000$ rows), from both a knowledge discovery perspective, statistical rule quality evaluation, and a predictive perspective, evaluating TFM-learned rules as part of interpretable rule-based classifiers (Section \ref{sec:evaluation}). 
\end{enumerate}

\section{Preliminaries and Related Work}\label{sec:related-work}

\paragraph{Association Rule Mining} The ARM problem originally seeks rules of the form $X \rightarrow Y$ over categorical data that satisfy minimum \emph{support} and \emph{confidence} thresholds, where support is the proportion of instances in the data containing $X \cup Y$, and confidence is the conditional frequency of $Y$ among instances containing $X$~\citep{agrawal1994fast}.  

\paragraph{Rule explosion.} The ARM literature has been dominated by algorithmic methods, most notably the FP-Growth~\citep{han2000mining} algorithm that has many variations, including parallel FP-Growth~\citep{parallelfpgrowth} and FP-Growth on GPU~\citep{gpufpgrowth}. Such methods work in two stages: finding frequent itemsets of higher support and then forming rules of higher confidence based on given thresholds. However, the algorithmic methods suffer from rule explosion, that is, an excessive number of rules mined that are hard to interpret and post-process~\citep{moens2013frequent}. This is due to the combinatorial growth in both the number of items, i.e., unique values, and their potential combinations into frequent itemsets and rules. The common approaches to overcome rule explosion in ARM include applying item constraints~\citep{srikant1997mining,baralis2012generalized,yin2022constraint}, mining only top-k rules~\citep{fournier2012mining,nguyen2018etarm}, and closed itemset mining~\citep{zaki2002charm}, which refers to mining only frequent itemsets without frequent supersets of equal support. Beyond such search space reduction methods, neurosymbolic alternatives leverage compact learned data representations.

\paragraph{Neurosymbolic methods in ARM.} Despite the wide-scale success of deep learning, neural networks have only recently applied to ARM. Aerial+~\citep{aerial_plus} uses an undercomplete denoising autoencoder to learn low-dimensional data representations and reconstruct the original inputs. Association rules are then derived based on reconstruction success, with the undercomplete structure favoring statistically significant rules over exhaustive rule mining. There have been other neural network-based approaches to ARM~\citep{berteloot2024association,patel2022innovative}. To the best of our knowledge, Aerial+ is the \textit{only method} that explicitly addresses rule explosion. However, earlier work showed that in the small data regime, Aerial+ suffers from a lack of training data, leading to rules that do not accurately capture patterns in the data~\citep{aerial_tabpfn}.

\paragraph{Tabular foundation models.} Prior-data fitted networks (PFNs)~\citep{pfns} pioneered training neural models on large collections of synthetic tabular tasks to enable in-context generalization and serve as a precursor to tabular foundation models. Tabular foundation models (TFMs) extend this paradigm by pretraining large-scale architectures on synthetic or real tabular data to learn transferable representations~\citep{tabpfn-nature}. Representative TFMs include TabPFN~\citep{tabpfn-nature}, trained on vast synthetic priors, TabICL~\citep{qu2025tabicl}, which broadens in-context learning across larger tables, and TabDPT~\citep{ma2024tabdpt}, which leverages large-scale real tabular data. Despite the spreading success of TFMs across tabular tasks, application of TFMs to ARM has been limited to fine-tuning Aerial+ with embeddings from TabPFN~\citep{aerial_tabpfn}.


\paragraph{Our contribution.} While recent neurosymbolic methods address rule explosion using neural networks, they suffer from poor performance on small datasets, leading to rules that do not capture true patterns in the data. We introduce a model-agnostic framework that generalizes these approaches and enables TFMs, which excel in low-data regimes, to be applied to categorical ARM \textit{out-of-the-box} for the first time.


\section{Model-Agnostic Rule Learning Framework}
\label{sec:framework}

Let $I=\{i_1, i_2, \ldots, i_m\}$ be a finite set of $m$ items, and $D = \{t_1, t_2, \ldots, t_n\}$ a dataset of transactions, where each transaction $t_i \subseteq I$. For tabular data with $k$ categorical features $F = \{f_1, \ldots, f_k\}$, each feature $f_j$ admits a finite set of $c_j$ categories 
$\{f_j^1, \dots, f_j^{c_j}\}$. The item universe is then
\begin{equation}
    I = \bigcup_{j=1}^k \left\lbrace f_j^1, \dots, f_j^{c_j}\right\rbrace,
\end{equation}
and the number of items is $m = \sum_{j=1}^k c_j$. Each transaction can be represented as a one-hot encoded vector 
$\mathbf{x} \in \{0,1\}^m$ satisfying 
$\sum_{i=1}^{c_j} x[f_j^i] = 1$ for all $f_j \in F$.

\paragraph{Example.}
Table~\ref{tab:hepatitis_rules} shows a fragment of a clinical dataset where each patient record corresponds to a transaction containing items such as
\texttt{Sex=male}, \texttt{Antiviral=yes}, or \texttt{Class=LIVE}.
Rules such as
\[
\texttt{Antiviral=yes} \wedge \texttt{Malaise=yes} \rightarrow \texttt{Class=LIVE}
\]
capture interpretable associations between symptoms, treatments, and outcomes.

\begin{definition}[Association Rule]\label{def:association-rule}
An association rule is an implication $X \rightarrow Y$, where $X, Y \subseteq I$, is a first-order Horn clause with $|Y| = 1$, $|X| \geq 1$, and $X \cap Y = \varnothing$. The set $X$ is the \textit{antecedent} and $Y$ is the \textit{consequent}. 
\end{definition}

Classical rule \textit{mining} methods enumerate frequent itemsets and apply thresholding on support and confidence, often leading to a combinatorial explosion of rules. In contrast, we consider association rule \emph{learning} from trained probabilistic models that capture statistical regularities in the data.

\begin{definition}[Conditional Probabilistic Model]
Let $\mathcal{M}_\theta$ be a trained model that allows computing conditional queries of
the form
\begin{equation}
P_\theta(i \mid X), \quad i \in I, \; X \subseteq I,    
\end{equation}
interpreted as the probability estimated by the model that item $i$ holds given that all items in $X$ are observed to hold.
\end{definition}

\noindent
\textbf{Example.} Given the antecedent $X = \{\texttt{Antiviral=yes},$ $\texttt{Malaise=yes}\}$ from
Table~\ref{tab:hepatitis_rules}, the model $\mathcal{M}_\theta$ can be queried for
$P_\theta(\texttt{Class=LIVE} \mid X)$.

We \textit{formalize} association rule learning from such models $\mathcal{M}_\theta$ via two requirements.

\begin{requirement}[Antecedent Validation]
\label{req:antecedent}
A model $\mathcal{M}_\theta$ provides a scoring function $s_\theta : 2^I \to [0,1]$, where $s_\theta(X)$ quantifies how plausible the partial configuration $X$ is under the learned data distribution. An antecedent $X$ is considered valid if $s_\theta(X) \ge \tau_a$, for a user-specified threshold $\tau_a \in [0,1]$.
\end{requirement}

\noindent
Intuitively, this prevents the extraction of rules based on highly unlikely or
spurious combinations of feature values (e.g., rare co-occurrences in the
clinical data).

\begin{requirement}[Consequent Extraction]
\label{req:consequent}
Given a valid antecedent $X$, the model allows computing conditional probabilities 
$P_\theta(i \mid X)$ for all $i \in I \setminus X$.  
An item $i$ is accepted as a consequent of $X$ if $P_\theta(i \mid X) \ge \tau_c$, for a user-specified threshold $\tau_c \in [0,1]$.
\end{requirement}

\paragraph{Example.} Let $X = \{\texttt{Antiviral=yes}, \texttt{Malaise=yes}\}$.
Suppose the model assigns score
$s_\theta(X) = 0.81$, and the antecedent threshold is set to $\tau_a = 0.6$,
so that $X$ is accepted as a valid antecedent.
Querying the model for conditional probabilities yields:
\[
P_\theta(\texttt{Class=LIVE} \mid X) = 0.92, 
\ P_\theta(\texttt{Class=DIE} \mid X) = 0.08.
\]
With a consequent threshold of $\tau_c = 0.8$, the rule
\[
\texttt{Antiviral=yes} \wedge \texttt{Malaise=yes}
\rightarrow \texttt{Class=LIVE}
\]
is therefore extracted by the framework.

\begin{definition}[Association Validity Function]
Given a model $\mathcal{M}_\theta$, the association validity function is defined as
\begin{equation}
    \mathcal{V}_\theta(X \rightarrow Y)
=
\mathbf{1}[s_\theta(X) \ge \tau_a]
\cdot
\min_{i \in Y} P_\theta(i \mid X),
\end{equation}
where $X,Y \subseteq I$ and $\mathbf{1}[\cdot]$ is an indicator function equal to one if its argument is true, and 0 otherwise.
\end{definition}

\begin{definition}[Association Rule Learning]
Given a trained model $\mathcal{M}_\theta$ and thresholds 
$\tau_a, \tau_c \in [0,1]$, the association rule learning problem is to identify
the set of rules
\begin{equation}
    \mathcal{R}
=
\{ X \rightarrow Y \mid \mathcal{V}_\theta(X \rightarrow Y) \ge \tau_c \}.
\end{equation}
\end{definition}

\subsection{Two Paradigms for Framework Instantiation}
\label{sec:framework-init}

Requirements~\ref{req:antecedent} and~\ref{req:consequent} can be satisfied through different model architectures. The following are two examples of such \textit{neural} architectures. However, note that the model being a neural network is not a requirement.

\subsubsection{Multi-Target Rule Learning Paradigm}

We now show that multi-target models, such as categorical autoencoders, naturally instantiate the proposed framework. In this setting, a neural model is trained to reconstruct, predict, or decode a data instance from a possibly corrupted, masked, or noisy input. A \textit{multi-target instantiation} refers to any such model that simultaneously outputs predictions for all features and satisfies the framework requirements. Although not explicitly trained for rule learning, such models can be queried post hoc to extract rules by interpreting their outputs as conditional predictions under partial evidence.

\begin{definition}[Multi-Target Instantiation]
Let $\mathcal{M}_\theta : \mathbb{R}^m \to \mathbb{R}^m$ be a trained 
multi-target model that maps an input vector $\mathbf{x}$ to a reconstructed 
output $\hat{\mathbf{x}} = \mathcal{M}_\theta(\mathbf{x})$.  
We assume that the output is normalized per feature, such that for each feature 
$f_j \in F$, $\sum_{i \in f_j} \hat{\mathbf{x}}[i] = 1$, so that $\hat{\mathbf{x}}[i]$ can be interpreted as the predicted 
probability of item $i$.

Given an antecedent set $X \subseteq I$, we construct an input probe vector 
$\mathbf{x}(X) \in \mathbb{R}^m$ where $\mathbf{x}(X)[i'] = 1$ for all $i' \in X$, 
$\mathbf{x}(X)[j] = 0$ for all other items $j$ in features containing items from $X$, 
and $\mathbf{x}(X)[k] = 1/|f|$ for items $k$ in remaining features $f$. We define the induced conditional probability as
\begin{equation}
P_\theta(i \mid X) := \mathcal{M}_\theta(\mathbf{x}(X))[i], \forall i \in I.
\end{equation}
\end{definition}

\paragraph{Example.} Let $X = \{\texttt{Antiviral=yes}, \texttt{Malaise=yes}\}$. A sample input probe for $P_\theta(\texttt{Class=LIVE} \mid X)$ would be $p=x(X)=[1, 0, 1, 0, 0.5, 0.5]$, where $p[1], p[3] = 1$ \textit{marks} items in $X$, $p[2], p[4] = 0$ marks other items in features containing $X$, and $p[5], p[6]=0.5$ refers to remaining items. \textbf{Note} that other probing strategies are possible depending on model architecture, e.g., leaving $p[5], p[6]$ missing to indicate no prior knowledge instead of equal probabilities.

This instantiation satisfies the two requirements as follows.

\begin{itemize}
    \item \textbf{Antecedent validation (Req.~\ref{req:antecedent}).}  
    We define the antecedent score as $s_\theta(X) := \min_{i \in X} P_\theta(i \mid X).$

    \item \textbf{Consequent extraction (Req.~\ref{req:consequent}).}  
    For a valid antecedent $X$, any item $i \in I \setminus X$ is extracted as a consequent if $P_\theta(i \mid X) \ge \tau_c.$
\end{itemize}

\textbf{Aerial+}~\citep{aerial_plus} can be seen as one instantiation of this paradigm. It allows \textit{simultaneous} antecedent validation and consequent extraction in a single forward pass, as the entire item space is reconstructed at once.

\begin{algorithm}[t]
\caption{Multi-Target Rule Learning}
\label{alg:multi-target-paradigm}
\algtext*{EndFor}
\algtext*{EndIf}
\algtext*{EndWhile}
\begin{algorithmic}[1]
\Require Dataset $D$, model $\mathcal{M}_\theta$, thresholds $\tau_a, \tau_c$
\Ensure Rule set $\mathcal{R}$
\State $\mathcal{R} \leftarrow \emptyset$
\ForAll{antecedents of interest $X \subseteq I$}
    \State $\mathbf{x} \leftarrow \text{ConstructProbe}(X)$ 
    \State $\hat{\mathbf{x}} \leftarrow \mathcal{M}_\theta(\mathbf{x})$ 
    \If{$\hat{\mathbf{x}}[i'] \geq \tau_a, \forall i' \in X$}
        \State $Y \leftarrow \{i \in I \setminus X \mid \hat{\mathbf{x}}[i] \geq \tau_c\}$
        \State $\mathcal{R} \leftarrow \mathcal{R} \cup \{X \rightarrow y \mid y \in Y\}$
    \EndIf
\EndFor
\State \Return $\mathcal{R}$
\end{algorithmic}
\end{algorithm}

\textbf{Algorithm \ref{alg:multi-target-paradigm}} illustrates how such multi-target association rule learning works in practice. Lines 2-7 iterate over antecedents of interest, e.g., all 3 combinations of $i \in I$. Line 3 constructs an input probe of size $m$, as described earlier. Line 4 performs a forward pass over the model to obtain reconstruction probabilities per feature. Lines 5 and 6 validate antecedents and extract consequents using the thresholds $\tau_a$ and $\tau_c$, and line 7 stores association rules.

\subsubsection{Single-Target Rule Learning Paradigm}
\label{sec:algorithm}

We now consider predictive models that are trained under a feature-wise single 
prediction objective, \textit{single-target instantiation}, where each feature is predicted from the remaining ones. These models provide access to conditional predictions of the form 
$P_\theta(i \mid I \setminus f_j)$ for items $i$ belonging to feature $f_j$.  
By aggregating such predictions across all features, we obtain a vector 
$\hat{\mathbf{x}}$ that serves as an approximation of the full conditional 
prediction vector required by the framework.

\begin{definition}[Single-Target Instantiation]
\label{def:pred-based-inst}
Let $\mathcal{M}_\theta$ be a predictive model that, given an input vector 
$\mathbf{x}$ and a target feature index $j$, outputs a probability distribution
over the categories of feature $f_j$, $\mathcal{M}_\theta(\mathbf{x}, j) \in [0,1]^{|f_j|}.$ To obtain a prediction vector over all items, we construct
\begin{equation}
\hat{\mathbf{x}}[i] := \mathcal{M}_\theta(\mathbf{x}_{\setminus f_j}, j)[i],
\quad \text{for all } i \in f_j, \; j = 1,\dots,k,
\end{equation}

where $\mathbf{x}_{\setminus f_j}$ denotes the input with feature $f_j$ masked 
or removed to prevent information leakage.  
Aggregating these predictions over all features yields a vector 
$\hat{\mathbf{x}} \in [0,1]^m$ that can be interpreted as conditional 
probabilities over items.
\end{definition}

\paragraph{Satisfaction of framework requirements.}
This instantiation satisfies the two framework requirements as follows.

\begin{itemize}
    \item \textbf{Antecedent validation (Req.~\ref{req:antecedent}).}  
    As in the multi-target case, we define $s_\theta(X) := \min_{i \in X} \hat{\mathbf{x}}[i]$.

    \item \textbf{Consequent extraction (Req.~\ref{req:consequent}).}  
    For a valid antecedent $X$, $i \in I \setminus X$ is extracted as a consequent if $\hat{\mathbf{x}}[i] \ge \tau_c$.
\end{itemize}

\paragraph{Discussion.}
Unlike multi-target instantiation, single-target predictive models do not allow explicit 
\textit{marking} of antecedent features when predicting those same features, as this 
would introduce information leakage.  
As a result, single-target instantiations may provide weaker control over 
antecedent conditioning. We return to this limitation in Discussions in  
Section~\ref{sec:discussion}.

\textbf{TFMs} are a natural candidate for this paradigm as they excel at in-context prediction, accurately predicting features conditioned on others \textit{without re-training}, and are robust in low-data regimes, which are the main focus of this paper.

\begin{algorithm}[t]
    \caption{TabProbe: Rule Extraction from TFMs}
    \label{alg:tfm-narl}
    \algtext*{EndFor}
    \algtext*{EndIf}
    \algtext*{EndWhile}
    \begin{algorithmic}[1]
    \Require Dataset $D$, tabular foundation model $\mathcal{M}_\theta$, max antecedents $a$, thresholds $\tau_a, \tau_c$
    \Ensure Rule set $\mathcal{R}$
    \State $\mathcal{R} \leftarrow \emptyset$
    \State $F \leftarrow \text{GetFeatures}(D)$
    \For{$i \leftarrow 1$ to $a$}
        \ForAll{feature combination $S \in \binom{F}{i}$}
            \State $\mathbf{Q} \leftarrow \text{GenerateProbeVectors}(S, F)$
            \State $\hat{\mathbf{P}} \leftarrow \mathbf{0}^{|\mathbf{Q}| \times m}$
            \ForAll{feature $f_j \in F$}
                \State $\mathbf{X}_{\text{ctx}} \leftarrow \text{RemoveFeature}(D, f_j)$
                \State $\mathbf{y}_{\text{ctx}} \leftarrow \text{GetLabels}(D, f_j)$
                \State $\mathcal{M}_\theta.\text{fit}(\mathbf{X}_{\text{ctx}}, \mathbf{y}_{\text{ctx}})$
                \State $\mathbf{Q}^{-j} \leftarrow \text{RemoveFeature}(\mathbf{Q}, f_j)$
                \State $\hat{\mathbf{P}}[\cdot, f_j] \leftarrow \mathcal{M}_\theta.\text{predict\_proba}(\mathbf{Q}^{-j})$
            \EndFor
            \ForAll{probe $\mathbf{q}_r \in \mathbf{Q}$}
                \If{$\min_{i \in S} \hat{\mathbf{P}}[r, i] \geq \tau_a$} 
                    \ForAll{$i \in I \setminus S$ where $\hat{\mathbf{P}}[r, i] \geq \tau_c$}
                        \State $\mathcal{R} \leftarrow \mathcal{R} \cup \{S \rightarrow i\}$
                    \EndFor
                \EndIf
            \EndFor
        \EndFor
    \EndFor
    \State \Return $\mathcal{R}$
    \end{algorithmic}
\end{algorithm}

\begin{figure}[!b]
    \centering
    \includegraphics[width=0.7\linewidth]{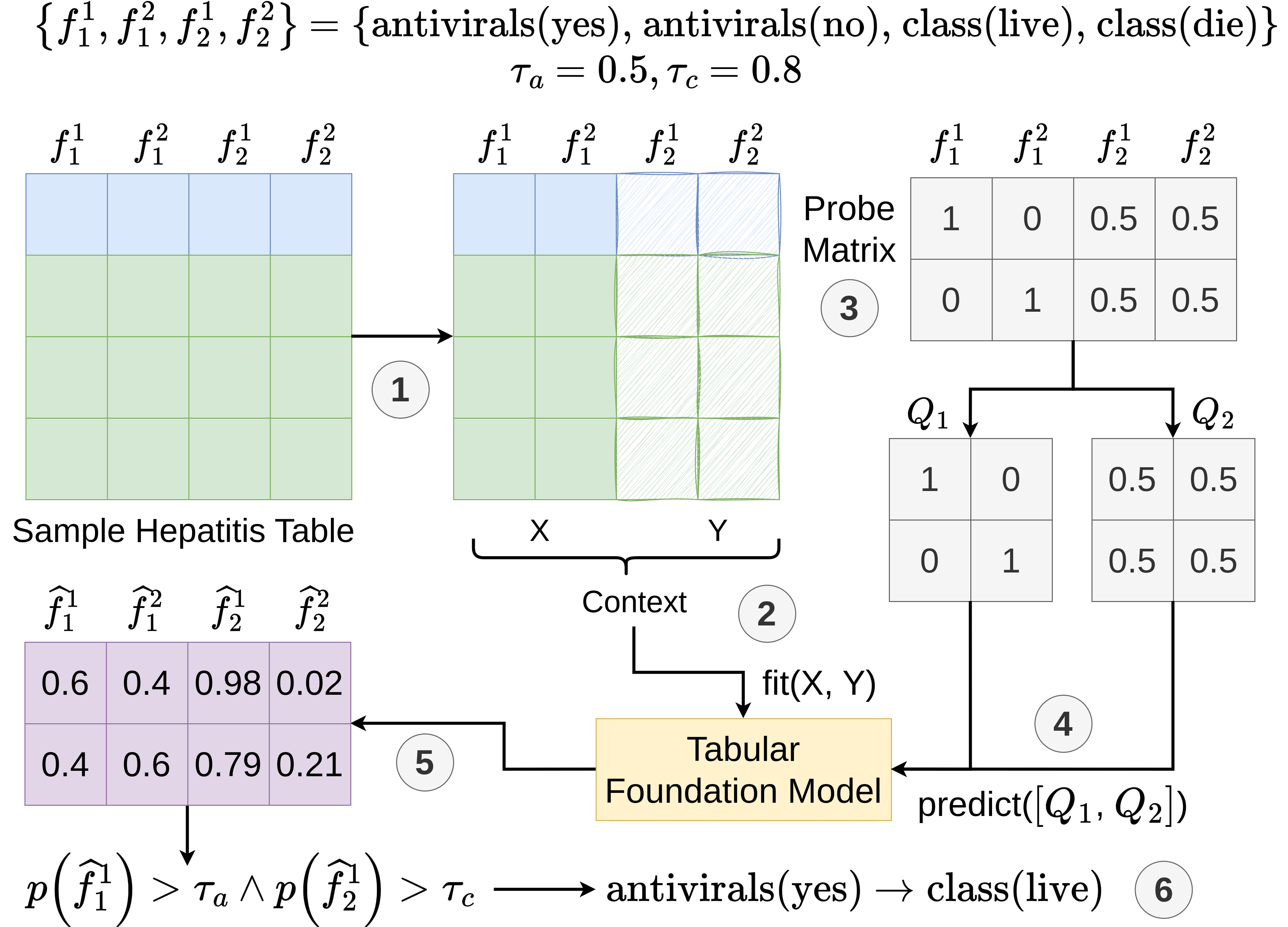}
    \caption{\textbf{Example} association rule learning from tabular foundation models using Algorithm \ref{alg:tfm-narl}: (1) split table into features (X) and labels (Y), (2) fit TFM on X and Y, (3) construct probing matrix with marked items (probability 1) and uniform priors for other items, (4) predict per feature, (5) aggregate predictions into reconstruction matrix, (6) apply thresholding to extract association rules.}
    \label{fig:example}
\end{figure}

\textbf{Algorithm \ref{alg:tfm-narl}} illustrates how the single-target association rule learning paradigm can be realized with TFMs. For a given max antecedent length of $a$ (i.e., rule complexity), lines 3-4 create item combinations for antecedents $S$, which are iterated over until line 16. Line 5, using \textit{GenerateProbeVectors()}, generates a probing vector $Q$ for marked features in $S$. Lines 7-12 iterate over each feature $f_j \in F$. \textit{RemoveFeature()} removes feature $f_j$ in D to prepare a context table (line 8). \textit{GetLabels()} extracts $f_j$ as context labels (line 9). Line 10 \textit{fits} the prepared context into the given TFM model $\mathcal{M}_\theta$ using in-context learning without training. \textit{RemoveFeature()} removes the feature $f_j$ from $Q$ (line 11), and line 12 performs a forward run on $\mathcal{M}_\theta$ with the \textit{predict\_proba()} function to obtain a probability distribution over values of $f_j$ stored in $\hat{P}$. Lines 13-16 iterate over the final prediction vector $\hat{P}$, perform antecedent validity (line 14) and consequent extraction (line 15) checks. Antecedent combinations and \textbf{probe vectors} are generated similarly to Algorithm \ref{alg:multi-target-paradigm}. Figure \ref{fig:example} illustrates a rule learning example from TFMs using Algorithm \ref{alg:tfm-narl}.

\paragraph{Time complexity.} Algorithm~\ref{alg:tfm-narl} has the time complexity $O(k^{a+1} \cdot (T_{\mathcal{M}_\theta}(n) + c_{\max}^a))$ where $k$ is the number of features, $a$ is the maximum antecedent size (typically 2-4), $T_{\mathcal{M}_\theta}$ is the underlying TFM architecture complexity, $n$ is the dataset size, and $c_{\max}$ is the maximum classes per feature. Please refer to \textbf{Appendix \ref{app:complexity-analysis}} for a detailed complexity analysis. Algorithm 1, however, is a factor of $k$ faster as it avoids per-feature predictions by reconstructing, prediction the entire data instance in a single forward pass.

\section{Evaluation}\label{sec:evaluation}

This section comprehensively evaluates TFMs as single-target association rule learners within 2 experimental settings; \textbf{i)} the \textbf{standard rule quality evaluation} procedure from the ARM literature, \textbf{ii)} \textbf{downstream classification evaluation} as part of interpretable machine learning models. Experiments are \textbf{repeated} for \textbf{10 random seeds} for robustness, and the average results are presented.

\textbf{Baselines.} We use 3 TFMs to learn association rules using Algorithm \ref{alg:tfm-narl}: TabPFNv2.5~\citep{tabpfnv2-5}, TabICL~\citep{qu2025tabicl}, and TabDPT~\citep{ma2024tabdpt}. Aerial+~\citep{aerial_plus} is taken as the state-of-the-art neurosymbolic baseline, representative of multi-target instantiation of our framework, and implemented using PyAerial~\citep{pyaerial}. FP-Growth~\citep{han2000mining} is taken as one of the most popular algorithmic ARM baselines, implemented with MLxtend~\citep{raschkas_2018_mlxtend}. \textbf{Note} that the output of any other algorithmic baseline is the same as FP-Growth, as they are all deterministic.

\textbf{Datasets.} 5 small ($\sim$100 rows) medical datasets, and 5 larger datasets (300 to 8000 rows) from the UCI ML repository~\citep{kelly2023uci} are used in all experiments. See \textbf{Appendix \ref{app:datasets}} for dataset metadata and preparation.

\textbf{Hyperparameters and fair comparison.} To ensure a fair comparison, each method is run with a max antecedent length of 2, TFMs, and Aerial+ is run with $\tau_c =0.8$, and FP-Growth with a min confidence threshold of $0.8$. In Setting 1, $\tau_a$ is set to 0.5 for TFMs and Aerial+, while FP-Growth is run with 3 different min support thresholds (0.3, 0.2, and 0.1), and all the results are reported. For Setting 2, $\tau_a = 0.3$ for TFMs and Aerial,+ and FP-Growth is run with 4 different min support thresholds (0.2, 0.1, 0.05, 0.01). Other parameters of Aerial+ are fine-tuned to obtain the best results possible. In Setting 2, on small tabular data, $\tau_a$ is reduced when necessary to obtain classification rules. All TFMs are run with entire tables as the context, and with an ensemble size of 8. See \textbf{Appendix \ref{app:hyperparameters}} for more details. 

\textbf{Source code} of all the experiments, including re-usable functions and the datasets are openly available at \href{https://github.com/DiTEC-project/tabprobe}{https://github.com/DiTEC-project/tabprobe}.

\phantomsection
\textbf{Hardware.} The experiments are run on a machine with an NVIDIA H100 GPU with 94~GiB memory, and a 4-core AMD EPYC 7F72 CPU per task.
\label{par:hardware}

\subsection{Setting 1: Rule Quality Evaluation}\label{sec:rule-quality-eval}

\paragraph{Setup.} In this experimental setting, we follow the standard association rule mining practice by computing statistical rule-quality metrics on the full dataset for rules learned by TFMs and baseline methods. Given a transaction dataset $D$ and a rule $X \rightarrow Y$, \textbf{rule quality metrics} are described below:

\begin{itemize}
    \item \textbf{Support:} Percentage of transactions containing all items in a given rule. Range [0, 1].

    \begin{equation}
        \textbf{support}(X \rightarrow Y) = \frac{|t \in (X \cup Y)|}{|D|}
    \end{equation}
    
    \item \textbf{Coverage:} Percentage of transactions covered by a given rule set. Range [0, 1].
    
    \begin{equation}
        \operatorname{coverage}(\mathcal{R}) = \frac{|\{\, t \in D \mid \exists\, (X \rightarrow Y) \in \mathcal{R} : X \subseteq t \,\}|}{|D|}
    \end{equation}
    
    \item \textbf{Confidence:} Percentage of transactions having $Y$ given $X$. Range [0, 1].

    \begin{equation}
        \text{confidence}(X \rightarrow Y) = \frac{\text{support}(X \rightarrow Y)}{\text{support}(Y)}
    \end{equation}
    
    \item \textbf{Zhang's metric}~\citep{zhangsmetric}\textbf{:} Association strength. Range [-1, 1]: -1 = dissociation, 0 = independence, and 1 = association.

    \begin{equation}
        \text{Zhang}(X \rightarrow Y) = \frac{\operatorname{confidence}(X \rightarrow Y) - \operatorname{confidence}(X' \rightarrow Y)}{\max\!\left(\operatorname{confidence}(X \rightarrow Y), \operatorname{confidence}(X' \rightarrow Y)\right)}
    \end{equation}

    where X' is the absence of X.
    
    \item \textbf{Interestingness:} How unexpected, surprising, is a rule in proportion to the support of its antecedent, consequent, and the transactions D. Range [0, 1].
    \begin{equation}
       \frac{\text{sup}(X \rightarrow Y)}{\text{sup}(X)} \cdot \frac{\text{sup}(X \rightarrow Y)}{\text{sup}(Y)} \cdot (1 - \frac{\text{sup}(X \rightarrow Y)}{|D|}) 
    \end{equation}
\end{itemize}

\textbf{Rule quality evaluation results.} Table \ref{tab:rule-quality} shows that TFMs can have full data coverage with up to 7 times fewer rules in comparison to the algorithmic methods (FP-Growth). This shows that TFMs can address rule explosion, similar to the Neurosymbolic baseline Aerial+. TabICL led to the highest association strength (Zhang) in both small and larger datasets, while also having the highest interestingness score in larger datasets, and the second highest in small datasets. FP-Growth with 0.3 min support leads to the highest interestingness and support score; however, with significantly lower data coverage, as FP-Growth requires lower support thresholds to have full data coverage, leading to rule explosion. Aerial+ resulted in significantly smaller data coverage, support, association strength, and interestingness than any other TFMs in small datasets. On larger datasets, Aerial+ was able to have higher data coverages, support, and interestingness, while still producing rules of significantly lower association strength than TFMs. The only metric that TFMs were consistently lower at is confidence; however, still being 85\% on average, which still refers to significant patterns in the data. See Appendix \ref{app:extended-results} for rule quality evaluation \textit{results per dataset}.

\begin{table}[!t]
    \centering
    \caption{Averages of rule quality metrics in percentage across 5 small and 5 larger datasets. Tabular foundation models can learn a concise number of rules with higher association strength, interestingness on average, and full data coverage, with slightly lower confidence scores (FP-G (x): refers to FP-Growth with x being min. support threshold).}
    \vspace{5pt}
    \label{tab:rule-quality}
    \begin{tabular}{lcccccc}
        \toprule
        \textbf{Algorithm} & \textbf{\# Rules} & \textbf{Support} & \textbf{Confidence} & \textbf{Zhang's metric} & \textbf{Interestingness} & \textbf{Coverage} \\
        \midrule
        \multicolumn{7}{c}{\emph{Small Tabular Datasets}} \\
        TabICL      & 378 & 38 & 84.3 & \textbf{31.8} & 42.6 & 98.1 \\
        TabPFNv2.5  & \textbf{286} & 38.9 & 86.3 & 30.3 & 43.2 & 98.1 \\
        TabDPT      & 294 & 36.6 & 82.9 & 26.2 & 40.9 & \textbf{1} \\
        Aerial+     & 313 & 28.7 & 89.3 & 23.5 & 34.4 & 88.1 \\
        FP-G (0.3)  & 430 & \textbf{42.4} & 89.8 & 23.6 & \textbf{48.6} & 99.7 \\
        FP-G (0.2)  & 680  & 34.5 & 90.9 & 24.4 & 41.1 & \textbf{1} \\
        FP-G (0.1)  & 1462 & 22.9 & \textbf{91.7} & 26 & 29.2 & \textbf{1} \\
        \midrule
        \multicolumn{7}{c}{\emph{Larger Tabular Datasets}} \\
        TabICL      & 11069 & 50.5 & 88.5 & \textbf{38.9} & \textbf{58.4} & 98.2 \\
        TabPFNv2.5  & 12238 & 48.8 & 86.6 & 34.3 & 55.8 & 99.0 \\
        TabDPT      & \textbf{5177} & 45.8 & 85 & 36.4 & 52.9 & 98.3 \\
        Aerial+     & 7214 & 50.9 & \textbf{93.6} & 22 & 57.5 & 97.6 \\
        FP-G (0.3)  & 26538 & \textbf{51} & 91.1 & 25.9 & 58 & 98.8 \\
        FP-G (0.2)  & 29211 & 43.5 & 91.4 & 25.2 & 50.4 & 99.6 \\
        FP-G (0.1)  & 35376 & 35.6 & 91.6 & 24.4 & 41.6 & \textbf{1} \\
        \bottomrule
    \end{tabular}
\end{table}

\textbf{Interpretation.} Rule quality evaluation leads to 3 core conclusions: i) TFMs can have full data coverage, even in low data settings, with a much smaller number of rules than exhaustive search, addressing rule explosion in ARM, ii) TFM-learned rules show stronger correlations (association strength, Zhang) and interestingness in both smaller and larger datasets. This implies strong non-redundancy in the final rule set, while the majority of the rules learned by algorithmic rule miners will be pruned out due to redundancy. iii) TFMs can address degraded performances in low-data scenarios, which affects the neurosymbolic baseline method Aerial+. \textbf{Why do TFMs show lower confidence but higher association strength?} TFMs achieve 3-6\% lower confidence than FP-Growth and Aerial+, while having up to 17\% higher association strength. We hypothesize this reflects learning generalizable patterns rather than dataset-specific correlations, as FP-Growth and Aerial+ can achieve high confidence by overfitting to biases in the training data that do not generalize to unseen data. This hypothesis is tested in experimental setting 2 using 5-fold cross-validation in downstream classification tasks.

\subsection{Setting 2: Downstream Task Evaluation}\label{sec:downstream-task-eval}

\textbf{Setup.} This experimental setting evaluates TFMs and baselines in a downstream classification task as part of two well-known rule-based interpretable ML models: CBA~\citep{cba,pyARC} and CORELS~\citep{corels}. CBA and CORELS receive a large set of rules (CBA) or frequent itemsets (CORELS), and select a subset to build a rule-based classifier for high-stakes decision-making. Algorithm \ref{alg:tfm-narl} is adapted to learn frequent itemsets to be able to run CORELS (see Appendix \ref{app:freq-items-tfms}). We first learn rules or itemsets with TFMs and the baselines, and then pass those to CBA and CORELS to build a classifier, and then measure accuracy, F1 score, precision, and recall. We use the same parameters as in Setting 1, except $\tau_a=0.3$ for Aerial+ and TFMs. The experiments are carried out for 10 random seeds and with \textit{5-fold cross-validation}, and the average results are presented. See Figure \ref{fig:rule-based-classifier} in Appendix \ref{app:extended-results} for an illustration of this setup.

\begin{table}[!t]
    \centering
    \caption{Predictive performance in percentage and rule (CBA) or itemset (CORELS) size given in column two. TFMs achieve better predictive performance with a concise number of rules in small tabular data, while being behind FP-G by only -1.5\% despite FP-G's rule explosion (FP-G (x) denotes FP-Growth with minimum support threshold $x$).}
    \vspace{5pt}
    \label{tab:predictive-performance}
    \begin{tabular}{lccccc}
        \toprule
        \textbf{Method} & \textbf{\# Rules} & \textbf{Accuracy} & \textbf{F1 score} & \textbf{Precision} & \textbf{Recall} \\
        \midrule
        \multicolumn{6}{c}{\emph{Small Tabular Datasets}} \\
        TabICL      & 616   & 82.93 & 81.07 & 81.13 & 82.93 \\
        TabDPT      & 641   & 82.22 & 80.35 & 80.58 & 82.22 \\
        TabPFNv2.5  & 657   & \textbf{84.47} & \textbf{82.75} & \textbf{83.02} & \textbf{84.47} \\
        Aerial+     & 579   & 80.69 & 77.35 & 76.51 & 80.69 \\
        FP-G (0.3)  & 266   & 79.72 & 76.45 & 75.84 & 79.72 \\
        FP-G (0.2)  & 422   & 79.95 & 77.08 & 76.86 & 79.95 \\
        FP-G (0.1)  & 934   & 83.54 & 81.37 & 82.49 & 83.54 \\
        FP-G (0.05) & 2066  & 83.08 & 80.63 & 81.42 & 83.08 \\
        FP-G (0.01) & 13991 & 83.28 & 80.58 & 81.16 & 83.28 \\
        \midrule
        \multicolumn{6}{c}{\emph{Larger Tabular Datasets}} \\
        TabPFNv2.5  & 8719  & 86.53 & 85.74 & 86.15 & 86.53 \\
        TabICL      & 8100  & 86.51 & 85.77 & 86.16 & 86.51 \\
        TabDPT      & 4341  & 85.37 & 84.67 & 85.22 & 85.37 \\
        Aerial+     & 9095  & 87.63 & 86.20 & 86.41 & 87.63 \\
        FP-G (0.3)  & 13454 & 84.90 & 84.36 & 84.53 & 84.90 \\
        FP-G (0.2)  & 14882 & 88.10 & \textbf{87.30} & \textbf{87.30} & 88.10 \\
        FP-G (0.1)  & 19588 & 87.87 & 86.50 & 86.33 & 87.87 \\
        FP-G (0.05) & 44424 & \textbf{88.13} & 86.61 & 86.34 & \textbf{88.13} \\
        FP-G (0.01) & 59603 & 88.04 & 86.60 & 86.41 & 88.04 \\
        \bottomrule
    \end{tabular}
\end{table}

\textbf{Results.} Table \ref{tab:predictive-performance} shows that TFMs, TabPFNv2.5 in particular, achieve better predictive performance with a concise number of rules in small tabular data. On larger datasets, Aerial+ outperforms TFMs by $\sim$1\% while FP-Growth outperforms TFMs by only $\sim$1.5\% despite having up to 10 times more rules. Both Aerial+ and TFMs led to a much smaller number of rules than algorithmic methods. See Appendix \ref{app:extended-results} for results per dataset and per classifier.

\textbf{Interpretation.} The results imply that both TFMs and Aerial+ can address rule explosion in ARM, as they learn a concise set of rules with better or comparable predictive performance than standard rule miners. Aerial+'s performance degrades in low-data scenarios while TFMs become the best option. Together with the rule quality evaluation results, we conclude that TFMs can address degraded performance in neurosymbolic approaches to ARM in low-data scenarios, both for knowledge discovery and interpretable high-stakes decision-making. Note that TFMs' performance can be further improved by reducing $\tau_a$ to obtain more rules of lower support (see Appendix \ref{app:hyperparameter-analysis} for \textit{hyper-parameter analysis}).

\subsection{Scalability Analysis} \label{sec:scalability}

\textbf{Setup.} TFMs and baselines are run for 10 random seeds for association rule learning with the exact same setup as Section \ref{sec:rule-quality-eval}, and the average execution times are recorded.

\begin{figure}[!t]
    \centering
    \includegraphics[width=0.7\linewidth]{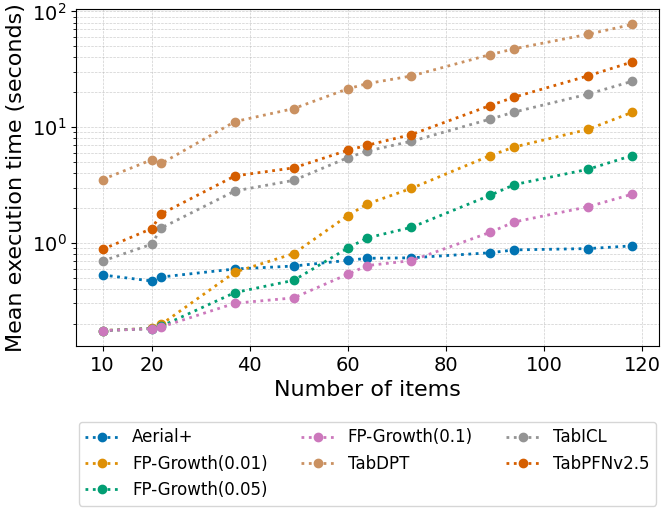}
    \caption{Mean execution time on an 8,124-rows dataset (mushroom) as itemset universe size increases. TFMs learn association rules within a minute on modest hardware for small-to-medium tables, whereas only Aerial+ scales to larger sizes (FP-Growth (x): x is min. support threshold).}
    \label{fig:exec-time}
\end{figure}

\textbf{Results.} Figure \ref{fig:exec-time} shows average execution times of TFMs and baselines when run on a table of 8,124-rows (mushroom) as the itemset universe increases, $I$ in Definition \ref{def:association-rule}. The results show that TFMs can learn association rules under a minute with up to 120 itemset universe size, equivalent to small-to-medium size tables, on affordable resources (see \nameref{par:hardware}). However, the TFMs are the slowest among all other methods, Aerial+ being the only method scalable on larger datasets.

\textbf{Interpretation.} This experiment shows that the current family of TFMs are fast enough to learn association rules on small-to-medium-sized tables within an acceptable time and on affordable resources, while they are not yet scalable enough to run on large tables. \textbf{Note} that our approach is applicable independently of the underlying TFM architectures. Considering the speed of advancement in TFMs, going from tables of up to 10,000 rows~\citep{hollmann2022tabpfn,tabpfn-nature} to 500,000 rows~\citep{ma2024tabdpt} within 2 years, we foresee that TFMs will be scalable to larger tables for association rule learning in the near future. Therefore, we recommend using TFMs for smaller-to-medium-sized tables and Aerial+ for larger tables, as it is more scalable and will benefit from larger sample sizes.

\section{Discussion}\label{sec:discussion}

\textbf{Model-agnostic association rule learning.} The framework introduced in Section \ref{sec:framework} applies to any conditional probabilistic model on tabular data that satisfies two requirements: antecedent validation and consequent extraction. Many neural models trained on tabular data can meet these requirements by adapting their outputs to probability distributions (e.g., via per-feature softmax) and therefore can be used for association rule learning. This framework offers a new avenue for neural network interpretability, as association rules provide inherently interpretable explanations of learned patterns over the data.

\textbf{Tabular foundation models can learn association rules.} We showed that TFMs with a classification objective can be adapted to learn association rules using the single-target instantiation of our framework out-of-the-box without re-training. TFMs led to a concise set of rules with high or full data coverage, and thus address rule explosion in ARM literature in both small and larger tabular data. Furthermore, TFM-learned rules showed stronger generalizability than the state-of-the-art, as they led to higher association strength and better downstream task performance on unseen data with cross-validation. In contrast to Aerial+, TFMs do not need training and architecture optimizations. Lastly, the hyper-parameter analysis in Appendix \ref{app:hyperparameter-analysis} shows that the antecedent validation and consequent extraction thresholds allow practitioners to balance rule set size and rule quality according to their needs.

\textbf{Scalability and limitations of tabular foundation models for rule learning.} The current family of TFMs is limited by context size, however, still reaching up to 500K rows~\citep{qu2025tabicl} on affordable resources. This limit, despite being significantly higher than the earlier TFMs (e.g., TabPFN v1), can prevent learning rules in very large tables. A second limitation is to have single-feature predictions at a time, slowing down rule learning by a factor of $k$ for $k$ features, as it requires $k$ predictions to obtain a final probability matrix (see Section \ref{sec:framework-init}, antecedent validation paragraph under Definition \ref{def:pred-based-inst}). TFMs use a frozen pretrained architecture, whereas neurosymbolic methods such as Aerial+ can optimize their architecture per dataset. This trade-off favors TFMs for small-to-medium datasets without any re-training according to our experimental results, but may favor optimized architectures like Aerial+ on very large datasets.  

\textbf{Practical implications and overcoming limitations.} Feature-wise predictions in TFMs can be parallelized across features to reduce execution time significantly. Some TFMs (e.g., TabPFNv2.5) support caching of context tables to speed up inference. Many concepts from ARM literature that reduce search space to address rule explosion can be applied to our framework and Algorithms \ref{alg:multi-target-paradigm} and \ref{alg:tfm-narl}. These include ARM with item constraints~\citep{yin2022constraint}, where we can create probing vectors for features of interest only, or top-k rule learning~\citep{fournier2012mining}, in which we learn rules of top-k highest probability rather than all. Another strategy is to construct probing vectors iteratively for features exceeding $\tau_a$ in the previous iteration only. This limits the probing matrix size and reduces computational costs.

\section{Conclusions and Future Work}

This paper introduced a model-agnostic association rule learning framework applicable to any conditional probabilistic model over tabular data satisfying two requirements: a scoring mechanism to validate antecedents and conditional probability estimation for consequent extraction. We instantiated the framework with tabular foundation models. In two comprehensive experimental settings, we showed that (i) tabular foundation models learn compact association rule sets with full data coverage, stronger associations, and less redundancy than state-of-the-art methods, addressing the degraded performance of neurosymbolic methods in low-data scenarios, and (ii) rules learned by tabular foundation models achieve equal or higher downstream classification performance as part of interpretable ML models.

Our framework opens multiple avenues for future work, including: (i) investigating other conditional probabilistic models for association rule learning beyond tabular foundation models and autoencoders; (ii) extending rule learning with TFMs (or other instantiations of our framework) to numerical data; (iii) learning rules from other data structures, such as graphs via graph neural network instantiations; and (iv) mitigating rule explosion in other branches of ARM, including sequential rule mining and high-utility itemset mining.

\textbf{Acknowledgment.} This work has received support from the Dutch Research Council (NWO), in the scope of the Digital Twin for Evolutionary Changes in water networks (DiTEC) project, file number 19454.

\newpage

\appendix

\section{Runtime Complexity Analysis}
\label{app:complexity-analysis}

This section provides a runtime complexity analysis of Algorithm~\ref{alg:tfm-narl} in big O notation.

\paragraph{Initials.} 
Building on the ARM problem definition given in Section \ref{sec:related-work} and Definition \ref{def:association-rule}; let $k$ denote the number of features in a tabular data, $c_{\max} = \max_j c_j$ the maximum categories per feature, $m = \sum_{j=1}^k c_j$ the total number of items, $n$ the number of transactions in $D$, $a$ the maximum antecedent size, and $T_{\mathcal{M}_\theta}(n)$ the time complexity of \textit{fitting} and \textit{predicting} with tabular foundation model $\mathcal{M}_\theta$ on $n$ rows.

\paragraph{Component-wise analysis.} 
We analyze each component of the algorithm separately and then aggregate the results.

\paragraph{Outer Loops.}
Lines 3-16 iterate over antecedent sizes 1 to $a$. Lines 4-16 generates $\binom{k}{i}$ feature combinations for each antecedent size $i$, and generates a total of 

\begin{equation}
\sum_{i=1}^{a} \binom{k}{i} = O(k^a) \quad \text{combinations for constant } a
\end{equation}

\paragraph{Probe Vector Generation.}
For each feature combination $S$ with $|S| = i$ in line 5, we generate probing vectors for all feature category combinations. The number of probing vectors is $\prod_{f \in S} c_f \leq c_{\max}^i$. Each probe has a dimension $m$. Therefore, generating each probe vectors $Q$ take $O(c_{\max}^i \cdot m)$.

\paragraph{Feature-wise Prediction.} Lines 7-12 iterates over the $k$ features and: 
\begin{itemize}
    \item Prepare $X_{ctx}$ context by removing feature $f_j$ from D in line 8, and storing it as $y_{ctx}$ in line 9: $O(1)$
    \item Fit the foundation model in line 10: $T_{\mathcal{M}_\theta}^{\text{fit}}(n)$
    \item Remove $f_j$ from $Q$ and predict on all remaining $Q^{-j}$ probing vectors: $T_{\mathcal{M}_\theta}^{\text{pred}}(c_{\max}^i)$
\end{itemize}
Total complexity of feature-wise predictions for $k$ features becomes: $O(k \cdot (T_{\mathcal{M}_\theta}(n) + T_{\mathcal{M}_\theta}(c_{\max}^i)))$

\paragraph{Rule Extraction.} Lines 13-16 extract rules from the final prediction matrix $\hat{P}$ by checking each $k$ features for $c^i_{max}$ probes in $O(k \cdot c^i_{max})$.

\paragraph{Total Complexity.}
Aggregating complexities across all components:

\begin{equation}
    \sum_{i=1}^{a} \binom{k}{i} \cdot [k \cdot (T_{\mathcal{M}_\theta}(n) + T_{\mathcal{M}_\theta}(c^i_{max})) + k \cdot c^i_{max}]
\end{equation}

Assuming $n \gg c_{max}$ for a typical tabular data and $a$ being constant in practice:

\begin{equation}
T_{\text{total}} = O\left( k^{a+1} \cdot T_{\mathcal{M}_\theta}(n) + k^{a+1} \cdot c_{\max}^a \right)
\end{equation}

Thus, Algorithm~\ref{alg:tfm-narl} has time complexity $O(k^{a+1} \cdot (T_{\mathcal{M}_\theta}(n) + c_{\max}^a))$ where $k$ is the number of features, $a$ is the maximum antecedent size (typically 2-4), $n$ is the dataset size, and $c_{\max}$ is the maximum classes per feature.

Compared to Aerial+, the single-target rule learning paradigm, instantiated with TFMs, requires $k$ forward passes per probe (one per feature), resulting in a factor of $k$ slowdown during rule extraction. However, when using pretrained foundation models, it eliminates the training phase, which can be advantageous for small datasets or rapid deployment.

\section{Extended Evaluation}\label{app:extended-evaluation}

This section includes an extended version of the \nameref{sec:evaluation} section with finer-grained results.

\subsection{Datasets}
\label{app:datasets}

10 datasets from the UCI ML repository~\citep{kelly2023uci} has been used in both of the experimental settings in Section \ref{sec:evaluation}. The dataset names together with number of columns, rows and unique items for ARM is shown in Table \ref{tab:datasets}. 

\begin{table}[t]
    \centering
    \caption{Small and larger tabular datasets from the UCI ML repository used in all the experimental settings.}
    \vspace{5pt}
    \label{tab:datasets}
    \begin{tabular}{lccc}
        \toprule
        \textbf{Dataset} & \textbf{Columns} & \textbf{Items} & \textbf{Rows} \\
        \midrule
        \multicolumn{4}{c}{\emph{Small Tabular Datasets}} \\
        Acute Inflammations & 6 & 22 & 120 \\
        Hepatitis & 19 & 89 & 155 \\
        Cervical Cancer & 19 & 197 & 72 \\
        ASD Screening (Adolescent) & 21 & 95 & 104 \\
        Fertility & 9 & 36 & 100 \\
        \midrule
        \multicolumn{4}{c}{\emph{Larger Tabular Dataset}} \\
        Breast Cancer & 9 & 43 & 286 \\
        Congressional Voting Records & 16 & 34 & 435 \\
        Mushroom & 22 & 118 & 8124 \\
        Chess (KR vs.\ KP) & 35 & 73 & 3196 \\
        Spambase & 57 & 157 & 4601 \\
        \bottomrule
    \end{tabular}
\end{table}

5 small medical tabular datasets ($\sim$100 rows) are used in evaluating TFMs in low-data scenarios, while 5 larger datasets (with 300 to 8000 rows) are used to emulate scenarios with enough data to train ML models. All of the datasets have a class label to evaluate TFMs in the downstream classification tasks as part of interpretable ML models (CBA~\citep{cba} and CORELS~\citep{corels}). Numerical columns are discretized using equal-frequency discretization with 10 folds. CSV version of the datasets can be found under \textit{data} folder of our repository \href{https://github.com/DiTEC-project/tabprobe}{https://github.com/DiTEC-project/tabprobe}.

\subsection{Parameter Settings}\label{app:hyperparameters}

\textbf{Fixed execution seeds for reproducibility.} All the experiments are repeated with the following 10 seeds, and the average results are presented. Seeds are listed in Table \ref{tab:seeds}.

\begin{table}[!b]
    \centering
    \caption{Fixed execution seeds used in all experimental settings for reproducibility of the results.}
    \vspace{5pt}
    \label{tab:seeds}
    \begin{tabular}{ll}
        \toprule
        \textbf{Seeds} & \textbf{Values} \\
        \midrule
        Seeds 1--4 & 42, 1608637542, 1273642419, 1935803228 \\
        Seeds 5--7 & 787846414, 996406378, 1201263687 \\
        Seeds 8--10 & 423734972, 415968276, 670094950 \\
        \bottomrule
    \end{tabular}
\end{table}

\textbf{TFM and baseline parameters.} Table \ref{tab:parameters} shows parameter settings for TFMs and baselines. For a fair comparison, maximum antecedent length is set to 2 for all methods, $\tau_c=0.8$ for TFMs and Aerial+, while the comparable minimum confidence threshold of FP-Growth is also set to $0.8$. $\tau_a = 0.5$ for TFMs and Aerial+ in experimental setting 1, while $\tau_a = 0.3$ in experimental setting 2 to obtain more classification rules. For fairness, we run FP-Growth with different minimum support thresholds and reported all the results; \{0.3, 0.2, 0.1\} for experimental setting 1, and \{0.3, 0.2, 0.1, 0.05, 0.01\} for experimental setting 2. We stopped at 0.1 in experimental setting 2 since FP-Growth leads to rule explosion, while setting 2 was run with a down to 0.01 as the goal of learning classification rules only restricted the rule space enough to run FP-Growth in reasonable execution times.

\begin{table}[t]
    \centering
    \caption{Parameter settings for TFMs and baselines across experimental settings. $a$ is antecedent length. \textsuperscript{\textdagger}$\tau_a=0.5$ for Setting 1, $\tau_a=0.3$ for Setting 2. \textsuperscript{\textdaggerdbl}min\_supp $\in \{0.3, 0.2, 0.1\}$ for Setting 1, $\in \{0.3, 0.2, 0.1, 0.05, 0.01\}$ for Setting 2.}
    \vspace{5pt}
    \label{tab:parameters}
    \begin{tabular}{lc}
        \toprule
        \textbf{Method}  & \textbf{Parameters} \\
        \midrule
        TabPFNv2.5 & $a=2$, $\tau_a \in \{0.5, 0.3\}$\textsuperscript{\textdagger}, $\tau_c=0.8$ \\
        TabICL & $a=2$, $\tau_a \in \{0.5, 0.3\}$\textsuperscript{\textdagger}, $\tau_c=0.8$ \\
        TabDPT & $a=2$, $\tau_a \in \{0.5, 0.3\}$\textsuperscript{\textdagger}, $\tau_c=0.8$ \\
        Aerial+ & $a=2$, $\tau_a \in \{0.5, 0.3\}$\textsuperscript{\textdagger}, $\tau_c=0.8$ \\
        FP-Growth & $a=2$, min\_conf$=0.8$, min\_supp $\in \{0.3, 0.2, 0.1, 0.05, 0.01\}$\textsuperscript{\textdaggerdbl} \\
        \bottomrule
    \end{tabular}
\end{table}

In experimental setting 2, while running CBA for specific datasets, we further reduced $\tau_a$ of TFMs and Aerial+ to be able to obtain classification rules that had lower support values. These are \textit{cervical\_cancer\_behavior\_risk} with $\tau_a = 0.1$ and \textit{breast\_cancer} with $\tau_a = 0.1$ (Aerial+ and TabDPT) and $\tau_a=0.05$ (TabPFNv2.5 and TabICL).

Aerial+ utilizes an under-complete denoising Autoencoder to learn a compact set of rules. This architecture should be optimized per dataset. Using a grid search, we found the parameters in Table \ref{tab:aerial-parameters} to work the best for Aerial+. We used those in our experiments and kept others at default values.

\begin{table}[!b]
    \centering
    \caption{Optimized Aerial+ architecture and training parameters per dataset; all other parameters kept at default values.}
    \vspace{5pt}
    \label{tab:aerial-parameters}
    \begin{tabular}{lccc}
        \toprule
        \textbf{Dataset} & \textbf{Batches} & \textbf{Layer Dimensions} & \textbf{Epochs} \\
        \midrule
        \multicolumn{4}{c}{\textit{Small Tabular Datasets}} \\
        Acute Inflammations & 2 & 4 & 10 \\
        Hepatitis & 2 & 4 & 10 \\
        Cervical Cancer & 18 & 2 & 20 \\
        ASD Screen. (Adolescent) & 2 & 4 & 10 \\
        Fertility & 2 & 4 & 10 \\
        \midrule
        \multicolumn{4}{c}{\textit{Larger Tabular Datasets}} \\
        Breast Cancer & 2 & 4 & 2 \\
        Congressional Vot. Recs. & 4 & 2 & 2 \\
        Mushroom & 64 & 4 & 2 \\
        Chess (KR vs.\ KP) & 64 & 64 & 2 \\
        Spambase & 64 & 64 & 2 \\
        \bottomrule
    \end{tabular}
\end{table}

\subsection{Extended Experimental Results}\label{app:extended-results}

This section presents extended experimental results for experimental setting 1 (Section \ref{sec:rule-quality-eval}) and setting 2 (Section \ref{sec:downstream-task-eval}). Please refer to those sections for the definitions and metrics used in each setting.

\subsubsection{Extended Rule Quality Evaluation}

\begin{table}[!t]
    \centering
    \caption{Rule quality evaluation of TFMs and baselines on small tabular datasets. In small tabular data, algorithmic ARM (FP-G) still leads to rule explosion. TFMs and Aerial+ can address rule explosion while TFMs find rules of higher statistical significance (Zhang and Interestingness). FP-G(x) refers to FP-Growth with x being min support threshold.}
    \label{tab:rule-quality-eval-small-tables}
    \vspace{5pt}
    \begin{tabular}{lcccccc}
        \toprule
        \textbf{Algorithm} & \textbf{\# Rules} & \textbf{Support} & \textbf{Confidence} & \textbf{Zhang's metric} & \textbf{Interestingness} & \textbf{Coverage} \\
        \midrule
        \multicolumn{7}{c}{\emph{Acute Inflammations}} \\
        TabICL & 46.1 & 32.9 & 80.7 & 57.8 & 47.3 & \textbf{1} \\
        TabPFNv2.5 & 51.9 & 31.8 & 81.2 & 60.9 & 46.7 & \textbf{1} \\
        TabDPT & 52.5 & 32.4 & 80.8 & 60.8 & 47.5 & \textbf{1} \\
        Aerial+ & 138.2 & 25.1 & 87.7 & 46.3 & 39.4 & \textbf{1} \\
        FP-G(0.3) & 46 & 37.7 & 92.4 & \textbf{62.1} & \textbf{58.8} & \textbf{1} \\
        FP-G(0.2) & 75 & 32.7 & 95.3 & 60.6 & 53.7 & \textbf{1} \\
        FP-G(0.1) & 123 & 25.6 & \textbf{96.1} & 55.1 & 42.6 & \textbf{1} \\
        \midrule
        \multicolumn{7}{c}{\emph{Hepatitis}} \\
        TabICL & 559.3 & 41.7 & 82 & 16.2 & 43.6 & \textbf{1} \\
        TabPFNv2.5 & 604.6 & 42 & 82.6 & 16.4 & 43.9 & \textbf{1} \\
        TabDPT & 565.6 & 42.1 & 82.1 & 15.2 & 44 & \textbf{1} \\
        Aerial+ & 677 & 37.1 & 87.2 & 16.4 & 40.9 & \textbf{1} \\
        FP-G(0.3) & 835 & 45.9 & 88.7 & \textbf{19.4} & \textbf{49.7} & \textbf{1} \\
        FP-G(0.2) & 1153 & 40.2 & 88.9 & 18 & 43.8 & \textbf{1} \\
        FP-G(0.1) & 2340 & 26.5 & \textbf{89.9} & 18.3 & 29.6 & \textbf{1} \\
        \midrule
        \multicolumn{7}{c}{\emph{Fertility}} \\
        TabICL & 50 & 36.9 & 85.4 & \textbf{11.0} & 37.4 & \textbf{1} \\
        TabPFNv2.5 & 42.9 & 39.1 & 87.9 & 7.8 & 39.2 & \textbf{1} \\
        TabDPT & 42.2 & 39.7 & 87.7 & 7.5 & 39.8 & \textbf{1} \\
        Aerial+ & 73.4 & 31.6 & 86.7 & 8.5 & 33.1 & \textbf{1} \\
        FP-G(0.3) & 52 & 41.8 & 88.5 & 3.6 & \textbf{42.1} & \textbf{1} \\
        FP-G(0.2) & 109 & 32.8 & 88.5 & 5.9 & 33.7 & \textbf{1} \\
        FP-G(0.1) & 287 & 20.5 & \textbf{89.3} & 10.3 & 22 & \textbf{1} \\
        \midrule
        \multicolumn{7}{c}{\emph{Cervical Cancer Behavior Risk}} \\
        TabICL & 14.5 & 41.9 & \textbf{92.9} & \textbf{58.6} & \textbf{46.1} & 90.3 \\
        TabPFNv2.5 & 13.8 & 38.6 & 92.1 & 48.1 & 41.9 & 90.4 \\
        TabDPT & 37.6 & 29.6 & 80.7 & 38.4 & 33.2 & 98.5 \\
        Aerial+ & 13.6 & 15.9 & 95.3 & 32 & 21.5 & 40.4 \\
        FP-G(0.3) & 27 & 40.4 & 90.7 & 18.4 & 43.9 & 98.6 \\
        FP-G(0.2) & 83 & 29.2 & 92 & 24.3 & 34.3 & \textbf{1} \\
        FP-G(0.1) & 476 & 16.6 & 92.7 & 31.6 & 23.8 & \textbf{1} \\
        \midrule
        \multicolumn{7}{c}{\emph{Autism Screening Adolescent}} \\
        TabICL & 1224.2 & 36.5 & 80.8 & 15.3 & 38.5 & \textbf{1} \\
        TabPFNv2.5 & 717.5 & 42.7 & 87.6 & \textbf{18.2} & 44.5 & \textbf{1} \\
        TabDPT & 773.2 & 39 & 83.1 & 9.1 & 40.1 & \textbf{1} \\
        Aerial+ & 663.7 & 33.8 & 89.9 & 14.2 & 37 & \textbf{1} \\
        FP-G(0.3) & 1193 & 46.3 & 88.9 & 14.8 & \textbf{48.8} & \textbf{1} \\
        FP-G(0.2) & 1982 & 37.5 & 89.5 & 13.5 & 39.8 & \textbf{1} \\
        FP-G(0.1) & 4088 & 25.5 & \textbf{90.2} & 14.8 & 28 & \textbf{1} \\
        \bottomrule
    \end{tabular}
\end{table}

Table \ref{tab:rule-quality-eval-small-tables} shows rule quality evaluation results for TFMs and baselines on small tabular datasets. The results show that even small tabular data may lead to rule explosion, as FP-Growth resulted in 4088 rules when run on the Autism Screening dataset with 10\% min support threshold. Increasing the min support threshold reduces the number of rules significantly; however, it also reduces the association strength of the rules learned. TFMs and Aerial+, however, can address rule explosion as they lead to full data coverage 4 over 5 datasets, and close to 100\% in the Cervical Cancer Behavior Risk dataset. Aerial+ produced rules of lower association strength in 4 out of 5 datasets, except Hepatitis, where Aerial+ and TabPFNv2.5 have the same association strength. In terms of the interestingness of the rules, TFM-learned rules outperformed Aerial+ on all datasets. The confidence score of TFMs is consistently lower than the baselines. We argue, and experimentally showed in setting 2 (see Section \ref{sec:downstream-task-eval}), that this is due to TFM-learned rules capturing more generalizable patterns as opposed to dataset-specific idiosyncrasies. This is partially implied by higher association strength and interestingness scores in TFM rules compared to baselines. The results in setting 2 showed that TFM-learned rules outperform the baselines when applied to unseen data in classification tasks.

\begin{table}[!t]
    \centering
    \caption{Rule quality evaluation of TFMs and baselines on larger tabular datasets. TFMs can reach full data coverage with up to 12 times fewer rules (Spambase), addressing rule explosion, and lead to rules of higher statistical significance (Zhang and Interestingness). FP-G(x) refers to FP-Growth with x being min support threshold.}
    \vspace{5pt}
    \label{tab:rule-quality-eval-larger-tables}
    \begin{tabular}{lcccccc}
        \toprule
        \textbf{Algorithm} & \textbf{\# Rules} & \textbf{Support} & \textbf{Confidence} & \textbf{Zhang's metric} & \textbf{Interestingness} & \textbf{Coverage} \\
        \midrule
        \multicolumn{7}{c}{\emph{Breast Cancer}} \\
        TabICL & 24.8 & 49.5 & \textbf{88.7} & \textbf{60.4} & \textbf{58.2} & 91.2 \\
        TabPFNv2.5 & 27.9 & 47.4 & 87.9 & 57.4 & 55.6 & 95.2 \\
        TabDPT & 27 & 47.5 & 88.2 & 58.5 & 55.8 & 91.7 \\
        Aerial+ & 15.7 & 43.9 & 88.1 & 36.4 & 52.4 & 90.2 \\
        FP-G(0.3) & 51 & 44.4 & 87.8 & 32.1 & 51.4 & 94.4 \\
        FP-G(0.2) & 112 & 33.4 & 87.7 & 26.7 & 39.2 & 97.9 \\
        FP-G(0.1) & 208 & 24.5 & 87.3 & 23.4 & 29.1 & \textbf{1} \\
        \midrule
        \multicolumn{7}{c}{\emph{Congressional Voting Records}} \\
        TabICL & 243.3 & 36 & 90.6 & \textbf{86.0} & 59.7 & \textbf{1} \\
        TabPFNv2.5 & 354.1 & 33 & 88.2 & 81.1 & 54.9 & \textbf{1} \\
        TabDPT & 312.6 & 34.5 & 89.4 & 83.1 & 56.3 & \textbf{1} \\
        Aerial+ & 224 & 34.8 & \textbf{92.1} & 61.4 & 56.6 & 97.8 \\
        FP-G(0.3) & 1016 & 35.5 & 89.1 & 70.6 & \textbf{60.6} & \textbf{1} \\
        FP-G(0.2) & 2024 & 30.7 & 88.6 & 65 & 52.9 & \textbf{1} \\
        FP-G(0.1) & 2867 & 26 & 88.5 & 60.5 & 44.7 & \textbf{1} \\
        \midrule
        \multicolumn{7}{c}{\emph{Mushroom}} \\
        TabICL & 1571.7 & 33 & 78.5 & \textbf{33.1} & 38.5 & \textbf{1} \\
        TabPFNv2.5 & 2240.1 & 27.8 & 70.9 & 19.5 & 31.8 & \textbf{1} \\
        TabDPT & 1752.2 & 27.1 & 67.3 & 11.8 & 30.3 & \textbf{1} \\
        Aerial+ & 857.6 & 37.2 & \textbf{95.5} & 8.6 & 39.4 & \textbf{1} \\
        FP-G(0.3) & 1247 & 44.3 & 93.7 & 17.8 & \textbf{48.6} & \textbf{1} \\
        FP-G(0.2) & 3147 & 31.8 & 95 & 24.5 & 38.7 & \textbf{1} \\
        FP-G(0.1) & 6882 & 22.2 & 95.3 & 27.6 & 29.3 & \textbf{1} \\
        \midrule
        \multicolumn{7}{c}{\emph{Chess (KR vs.\ KP)}} \\
        TabICL & 14731.9 & 54.4 & 88.9 & \textbf{9.7} & 55.1 & \textbf{1} \\
        TabPFNv2.5 & 13682.7 & 57.6 & 91.1 & 8.6 & \textbf{58.3} & \textbf{1} \\
        TabDPT & 14454.4 & 56.4 & 90 & 8.7 & 57 & \textbf{1} \\
        Aerial+ & 11901.7 & 57.4 & \textbf{94.8} & 1.3 & 57.6 & \textbf{1} \\
        FP-G(0.3) & 21064 & 57.4 & 93.1 & 4.5 & 57.2 & \textbf{1} \\
        FP-G(0.2) & 27604 & 49.8 & 93.3 & 4.8 & 49.9 & \textbf{1} \\
        FP-G(0.1) & 35358 & 42.1 & 93.5 & 5.6 & 42.6 & \textbf{1} \\
        \midrule
        \multicolumn{7}{c}{\emph{Spambase}} \\
        TabICL & 38775.1 & 79.8 & 95.6 & 5.1 & 80.2 & \textbf{1} \\
        TabPFNv2.5 & 44888.5 & 78 & 95.1 & 5 & 78.5 & \textbf{1} \\
        TabDPT & 9340.8 & 63.6 & 90.1 & \textbf{20.0} & 65.2 & \textbf{1} \\
        Aerial+ & 23075.1 & 81.4 & \textbf{97.7} & 2.5 & \textbf{81.7} & \textbf{1} \\
        FP-G(0.3) & 109312 & 73.6 & 92.2 & 4.7 & 72.6 & \textbf{1} \\
        FP-G(0.2) & 113168 & 72 & 92.3 & 4.9 & 71.2 & \textbf{1} \\
        FP-G(0.1) & 131567 & 63.4 & 93.2 & 4.9 & 62.6 & \textbf{1} \\
        \bottomrule
    \end{tabular}
\end{table}

Table \ref{tab:rule-quality-eval-larger-tables} shows rule quality evaluation results for larger tabular datasets. On these datasets, the impact of rule explosion is more visible, e.g., as FP-Growth produces more than 100.000 rules on the Spambase dataset even with 30\% min support threshold. Both TFMs and Aerial+ were again able to learn much more compact rule sets with higher statistical significance (Zhang and interestingness) than FP-Growth. The confidence scores of TFMs were relatively higher in large datasets than in small datasets when compared to baselines. TFM-learned rules have the highest association strength on all datasets and full data coverage in 4 out of the 5 datasets. 

\subsubsection{Extended Downstream Task Evaluation}

This section presents extended downstream task evaluation results of TFMs and baselines per dataset (see Section \ref{sec:downstream-task-eval}). 

\begin{figure}[!t]
    \centering
    \caption{Experimental Setting 2 for downstream task evaluation of TFMs and baselines. (1) tabular data is split into training as test folds for cross validation, (2) TFMs and baselines are run on the training fold to learn rules (CBA) or frequent itemsets (CORELS work with frequent itemsets and not rules), (3) learned rules and itemsets are passed to CBA and CORELS to build a rule-based classifier, (4) rule-based classifiers are evaluated on the test fold. This process is repeated 5-fold across 10 random seeds.}
    \label{fig:rule-based-classifier}
    \vspace{5pt}
    \includegraphics[width=0.7\linewidth]{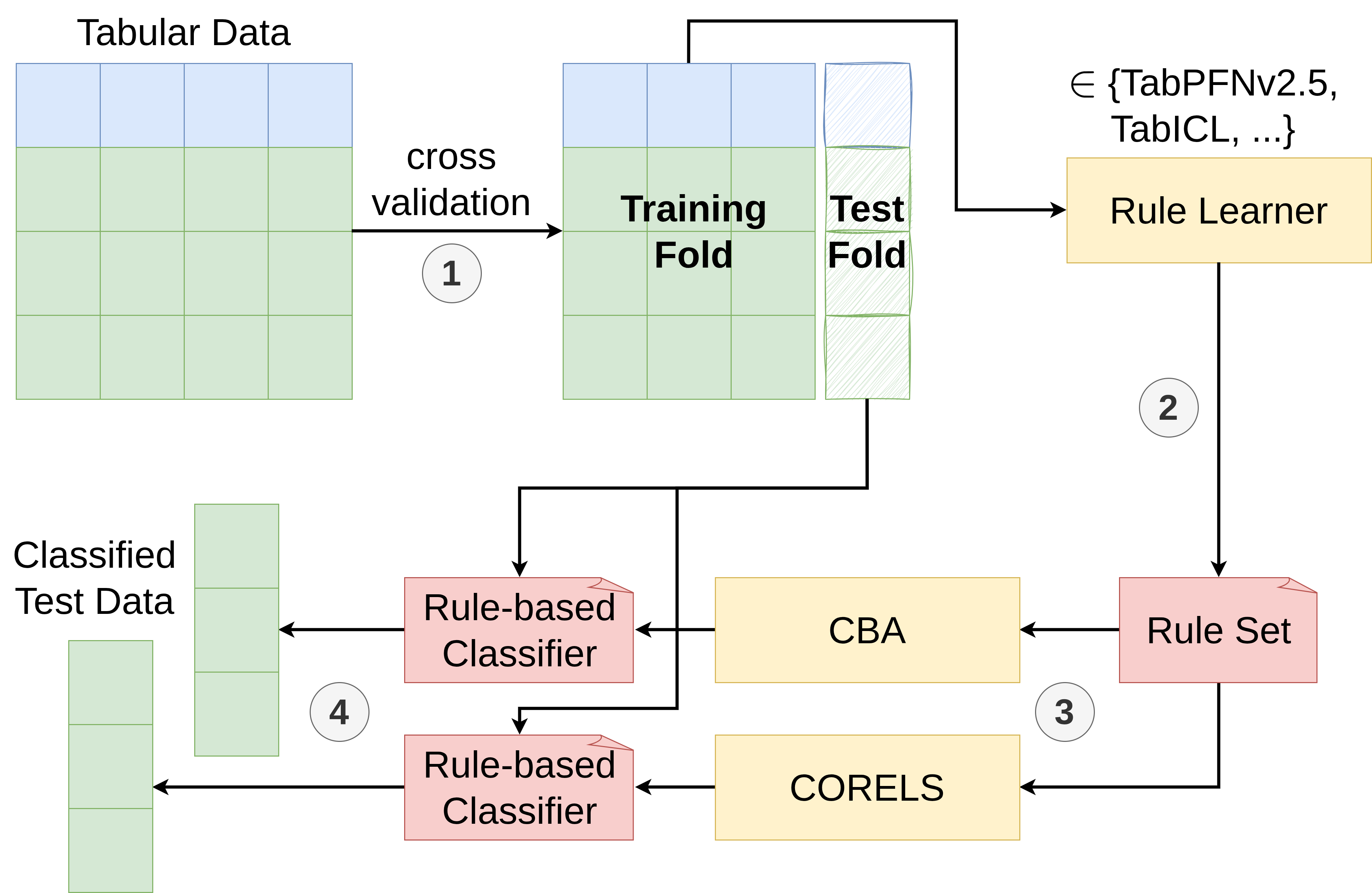}
\end{figure}

\textbf{Experimental setup} for downstream task evaluation is shown in Figure \ref{fig:rule-based-classifier}. CBA~\citep{cba} and CORELS~\citep{corels} are two well-known interpretable ML models used in high-stakes decision-making. CBA receives a set of association rules while CORELS receives a set of frequent itemsets to select and order a subset to build a rule-based classifier. Each tabular dataset is first split into training and test folds for 5-fold cross-validation. TFMs and baselines are run on the training folds to learn association rules for CBA and frequent itemsets for CORELS (see Section \ref{app:freq-items-tfms} for an adaptation of Algorithm \ref{alg:tfm-narl} to frequent itemset learning). Rule-based classifiers are then built using CBA and CORELS and evaluated on the test folds based on accuracy, precision, recall, and F1 score.

\begin{figure*}[h]
    \centering
    \includegraphics[width=\linewidth]{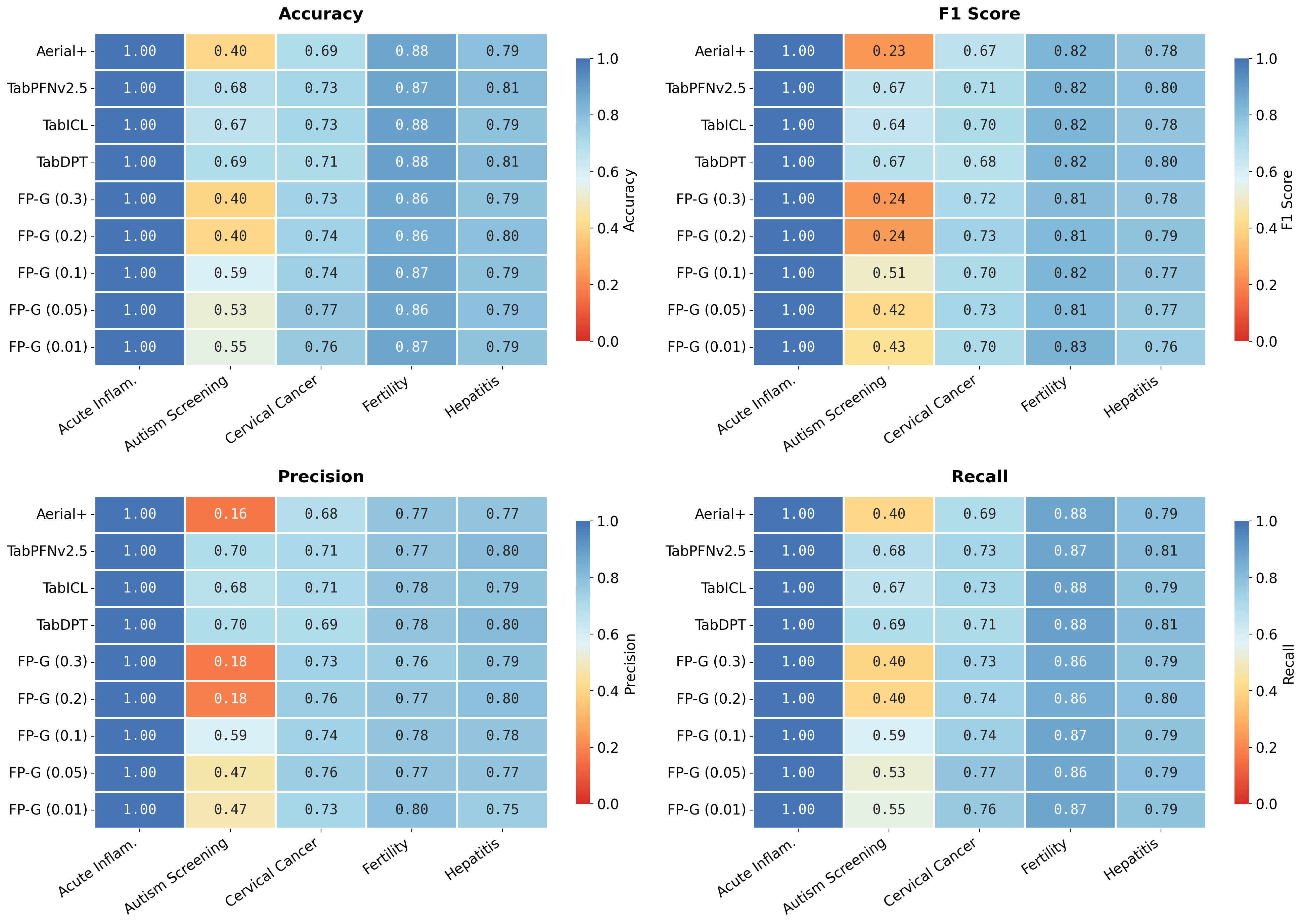}
    \caption{Classification tasks results with CORELS on small tabular data. TFMs have better or compatible ($\pm 1\%$) predictive performance on 4 out of 5 small datasets, except the Cervical Cancer dataset. This shows that the small number of rules that TFMs learn is indeed more generalizable than the baselines on unseen data.}
    \label{fig:corels-small}
\end{figure*}

\begin{figure*}[h]
    \centering
    \includegraphics[width=\linewidth]{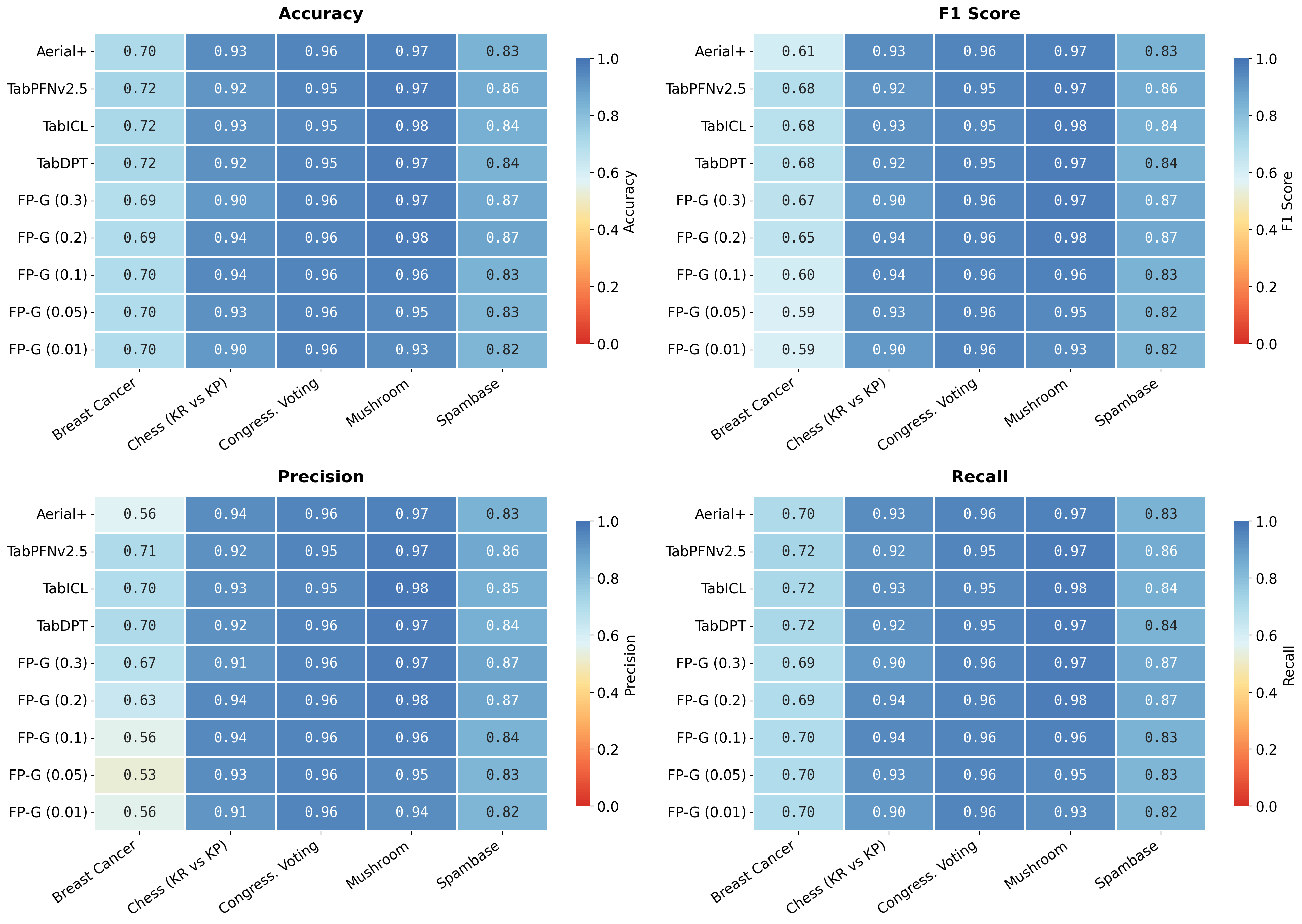}
    \caption{Classification task results with CORELS on large tabular data. On large datasets, TFMs have achieved better or comparable performance on all datasets compared to the baselines.}
    \label{fig:corels-large}
\end{figure*}

\begin{figure*}[h]
    \centering
    \includegraphics[width=\linewidth]{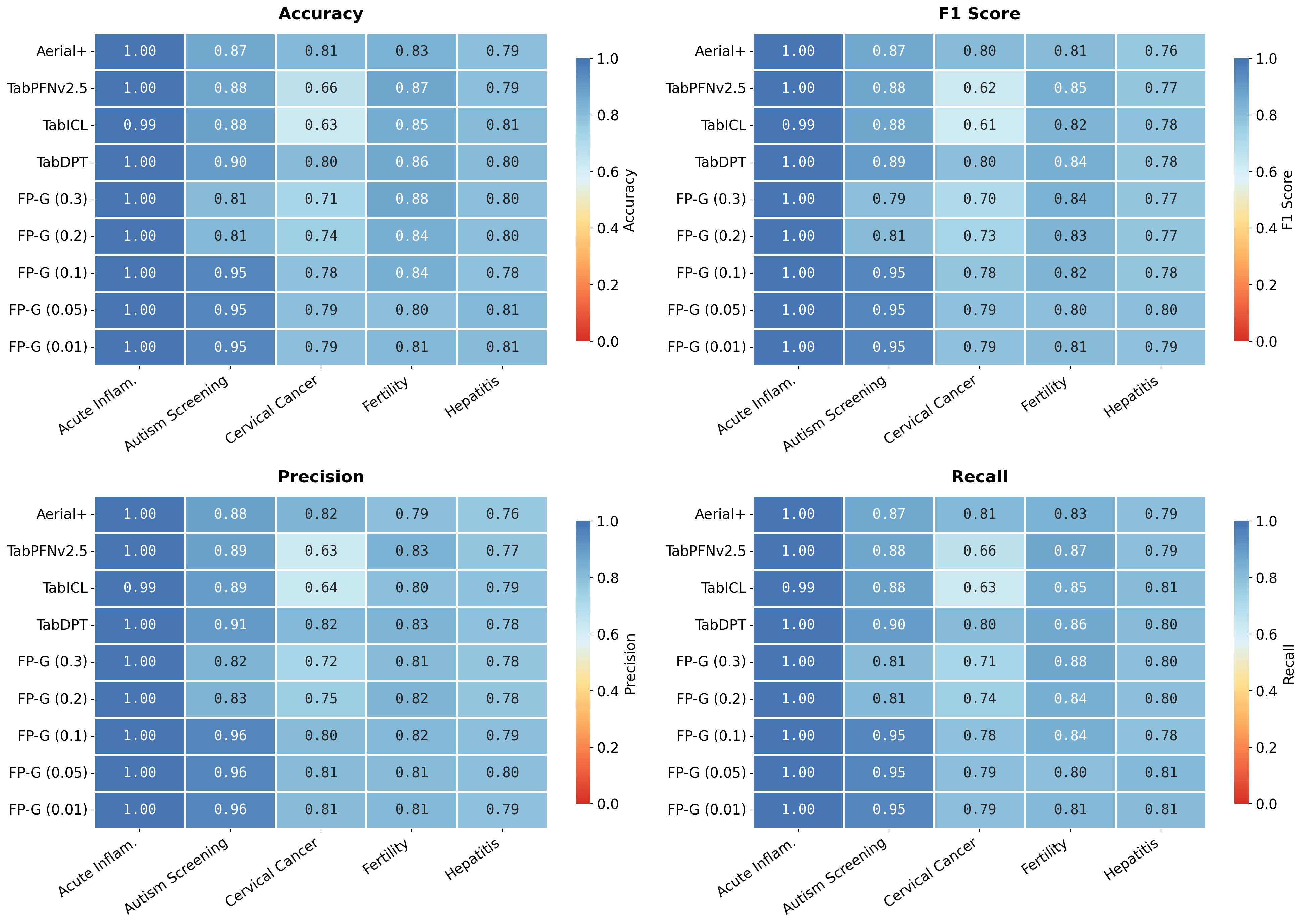}
    \caption{Classification tasks results with CBA on small tabular data. TFMs have better or compatible predictive performance than the baselines in 4 out of 5 datasets, except the Autism Screening dataset, while utilizing a small number of rules.}
    \label{fig:cba-small}
\end{figure*}

\begin{figure*}[h]
    \centering
    \includegraphics[width=\linewidth]{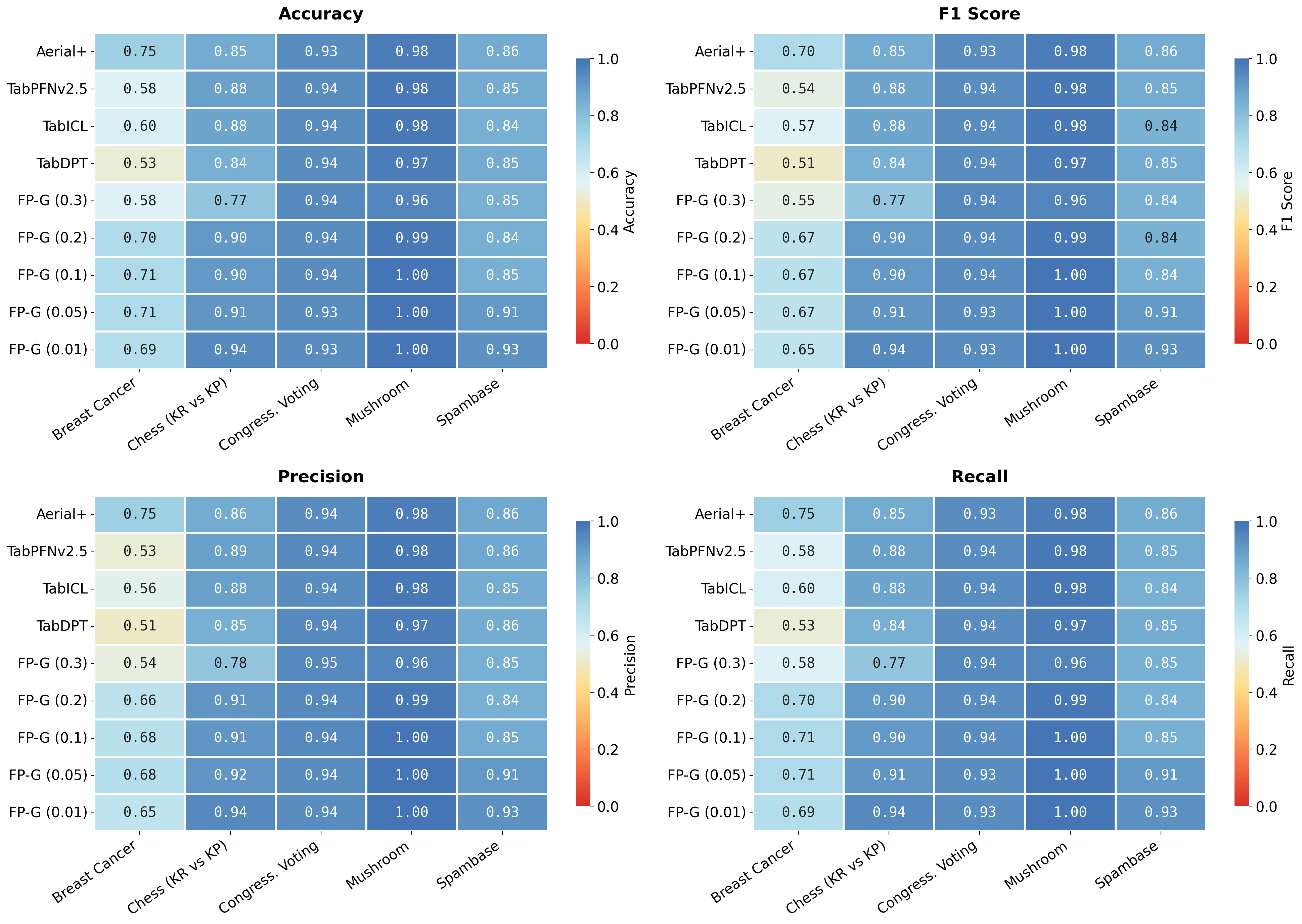}
    \caption{Classification tasks results with CBA on large tabular data. TFMs resulted in better or compatible predictive performance in only 2 out of the 5 large datasets with CBA. Note that CBA prioritizes rules of higher support and confidence rather than rules of higher association strength and generalizability.}
    \label{fig:cba-large}
\end{figure*}

\textbf{The results} of classification tasks for TFMs and baselines per dataset are given in \Cref{fig:corels-small,fig:corels-large,fig:cba-small,fig:cba-large}. With CORELS, TFMs have led to higher or compatible ($\pm2\%$) predictive performance on all large tabular datasets compared to the baselines, and 4 out of 5 of the small tabular datasets except the Cervical Cancer, where TabPFNv2.5 is behind $\sim2\%$ FP-Growth with 0.05 min support threshold. With CBA, TFMs again have led to higher or compatible scores on 4 out of 5 small tabular datasets compared to the baselines. On large datasets, however, TFMs were able to outperform baselines on 2 out of 5 datasets. 

\subsubsection{Interpretation of the Overall Evaluation Results}

Rule quality, downstream tasks, and scalability evaluations together imply the following 3 conclusions:

\begin{enumerate}
    \item \textbf{TFMs can address rule explosion in ARM.} Our approach can utilize TFMs out-of-the-box to learn a concise set of association rules with high data coverage and high statistical rule quality similar to Aerial+. 
    \item \textbf{TFMs remain robust in low-data scenarios for association rule learning.} While the performance of Aerial+ degrades on small tabular data in terms of association strength (Zhang), interestingness, and predictive performance, TFMs remain robust in such scenarios and outperform Aerial+ and FP-Growth in rule quality as well as in predictive performance.
    \item \textbf{TFMs-based rule learning can scale to medium-sized tables under a minute.} Section \ref{sec:scalability} showed that TFMs can scale up to itemset universe size of 120 with 8124 data instances, which corresponds to medium-sized tables, under a minute on affordable resources (see \nameref{par:hardware}). On the other hand, Aerial+ is the only method that can scale to larger datasets. Note that our method is independent of the underlying TFM. Considering the rapid developments in TFM capabilities within the past 2 years, we think that TFMs will be scalable to larger datasets in the near future. 
\end{enumerate}

\section{Hyperparameter Analysis}\label{app:hyperparameter-analysis}

This section analyzes the hyperparameters of Algorithm $\ref{alg:tfm-narl}$.

\textbf{Setup.} Each of the TFMs is run on all 10 datasets with 10 random seeds with varying $\tau_a$ and $\tau_c$ thresholds. First, $\tau_c = 0.8$ is fixed and $\tau_a \in \{0.1, 0.2, 0.3,0.4,0.5\}$ is varied. Next, $\tau_a = 0.5$ is fixed and $\tau_c \in \{0.5, 0.6, 0.7, 0.8, 0.9\}$ is varied. Algorithm \ref{alg:tfm-narl} is executed per TFM-dataset pair up until line 12 to obtain $\hat{P}$ (predicted probability distributions per feature). And then the aforementioned thresholds are applied to the same $\hat{P}$. This ensures measuring the impact of hyperparameters on an identical output matrix for comparable results. 

\begin{figure*}[h]
    \centering
    \includegraphics[width=\linewidth]{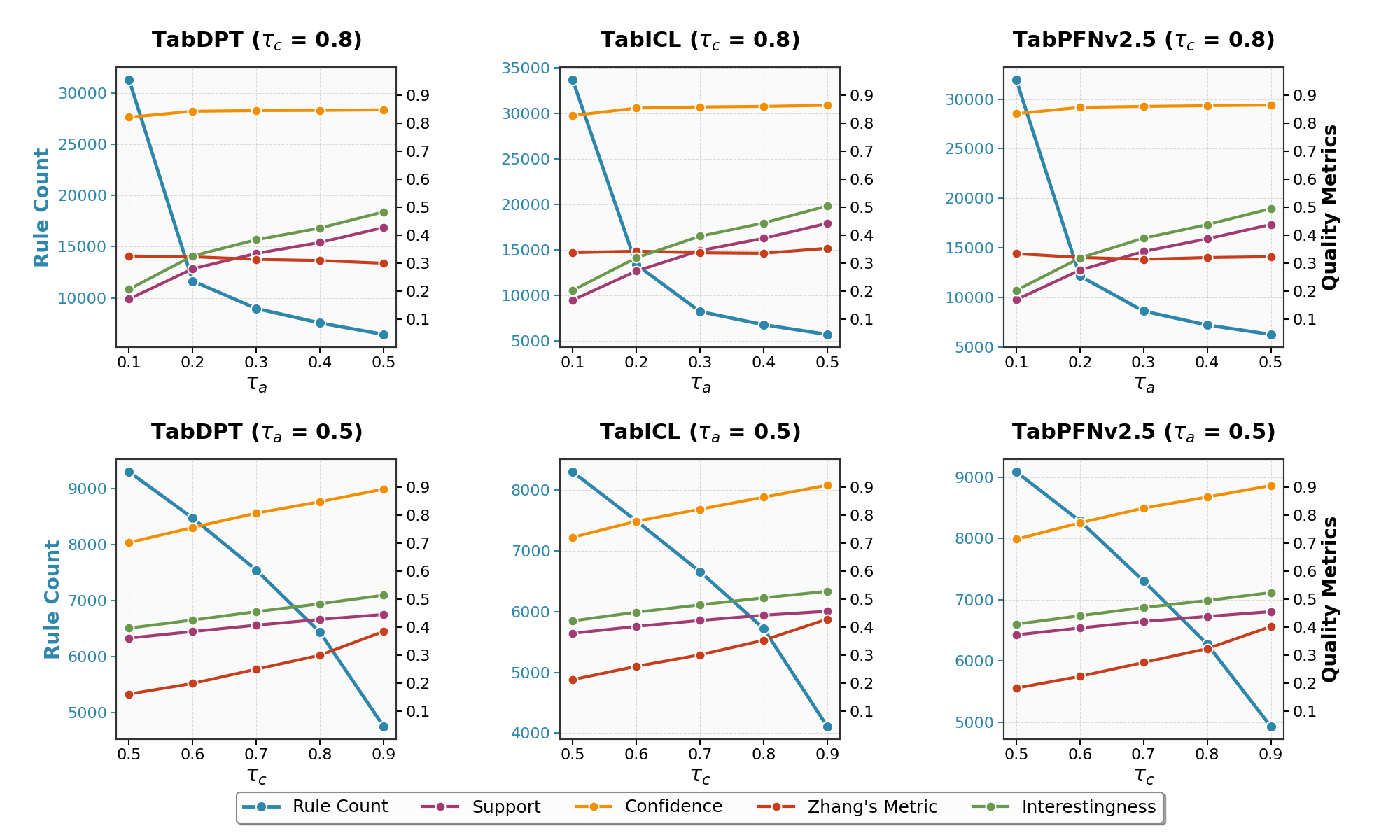}
    \caption{Hyperparameter analysis of Algorithm \ref{alg:tfm-narl} over all 10 datasets. The left y-axis specifies rule counts, and the right y-axis specifies average rule quality scores. The first row shows that increasing $\tau_a$ (while keeping $\tau_c = 0.8$ fixed) leads to a smaller number of rules, higher support and interestingness, while the association strength (Zhang) and confidence scores do not change significantly. The second row shows that increasing $\tau_c$ (while keeping $\tau_a = 0.5$ fixed) leads to higher support, confidence, association strength, and interestingness scores and reduces the number of rules learned.}
    \label{fig:hyperparam-analysis}
\end{figure*}

\textbf{The results} of this analysis is presented in Figure \ref{fig:hyperparam-analysis}. This figure shows the change in average rule counts and rule quality scores (support, confidence, Zhang's metric, and interestingness) as $\tau_a$ and $\tau_c$ change per TFM. The first row shows that as $\tau_a$ increases, rule count decreases while support and interestingness scores increase. No significant change is observed in confidence and association strength (Zhang) scores. The second row shows that as $\tau_c$ increases, support, confidence, association strength, and interestingness scores increase significantly, while rule count decreases.

\textbf{The implications} of the results are that $\tau_a$ impacts rule quality synonymously to the minimum support threshold of algorithmic methods, while $\tau_c$ has a more complex impact than the minimum confidence threshold. Hyperparameter settings depend on the rule learning task, as rules of different quality may lead to different results across tasks. As an example, in an anomaly detection task, when considered as a classification task, it can be carried out with a rule-based classifier that should include rules of lower support. Therefore, lower $\tau_a$ values should be preferred. On the contrary, in a knowledge discovery task with the goal of finding interesting patterns in the data, both $\tau_a$ and $\tau_c$ should be set sufficiently high to optimize rule interestingness and association strength, while avoiding an excessively small number of rules.

\section{Frequent Itemset Mining with Tabular Foundation Models}
\label{app:freq-items-tfms}

As part of experimental setting 2 (Section \ref{sec:downstream-task-eval}), we utilized CORELS~\citep{corels} to build rule-based classifiers for downstream task evaluation besides CBA for robust results. CORELS require frequent itemsets to build classifiers. Therefore, this section provides an adaptation of Algorithm \ref{alg:tfm-narl} to learn frequent itemsets with TFMs.

\begin{algorithm}[t]
    \caption{Frequent itemset mining with TFMs}
    \label{alg:tfm-freq-items}
    \algtext*{EndFor}
    \algtext*{EndIf}
    \algtext*{EndWhile}
    \begin{algorithmic}[1]
    \Require Dataset $D$, tabular foundation model $\mathcal{M}_\theta$, max antecedents $a$, \colorbox{green!20}{similarity threshold $\tau_s$}
    \Ensure Frequent itemsets $\mathcal{I}$
    \State \colorbox{green!20}{$\mathcal{I} \leftarrow \emptyset$}
    \State $F \leftarrow \text{GetFeatures}(D)$
    \For{$i \leftarrow 1$ to $a$}
        \ForAll{feature combination $S \in \binom{F}{i}$}
            \State $\mathbf{Q} \leftarrow \text{GenerateProbeVectors}(S, F)$
            \State $\hat{\mathbf{P}} \leftarrow \mathbf{0}^{|\mathbf{Q}| \times m}$
            \ForAll{\colorbox{green!20}{feature $s_j \in S$}}
                \State $\mathbf{X}_{\text{ctx}} \leftarrow \text{RemoveFeature}(D, f_j)$
                \State $\mathbf{y}_{\text{ctx}} \leftarrow \text{GetLabels}(D, f_j)$
                \State $\mathcal{M}_\theta.\text{fit}(\mathbf{X}_{\text{ctx}}, \mathbf{y}_{\text{ctx}})$
                \State $\mathbf{Q}^{-j} \leftarrow \text{RemoveFeature}(\mathbf{Q}, f_j)$
                \State $\hat{\mathbf{P}}[\cdot, f_j] \leftarrow \mathcal{M}_\theta.\text{predict\_proba}(\mathbf{Q}^{-j})$
            \EndFor
            \ForAll{probe $\mathbf{q}_r \in \mathbf{Q}$}
                \If{\colorbox{green!20}{$\min_{i \in S} \hat{\mathbf{P}}[r, i] \geq \tau_s$}}
                    \State \colorbox{green!20}{$\mathcal{I} \leftarrow \mathcal{I} \cup \{S\}$}
                \EndIf
            \EndFor
        \EndFor
    \EndFor
    \State \Return $\mathcal{I}$
    \end{algorithmic}
\end{algorithm}

The algorithm to learn frequent itemsets with TFMs out-of-the-box is given in Algorithm \ref{alg:tfm-freq-items}. Differences to Algorithm \ref{alg:tfm-narl} are marked with a \colorbox{green!20}{green} background color. Instead of two probabilistic antecedent and consequent thresholds, $\tau_a$ and $\tau_c$, we now have a single similarity threshold $\tau_s$. In the loop between lines 7-12, the algorithm queries TFMs per feature in $S$ only and not all the features in $F$. In lines 14-15, instead of threshold checks on antecedents and consequents separately, the $\tau_s$ threshold is checked against the marked features in $S$. The feature sets that have higher prediction probabilities than $\tau_s$ are then saved and returned as the frequent itemsets.

\end{document}